\documentclass[final]{cvpr}

\usepackage{diagbox}
\usepackage{overpic}
\usepackage{times}
\usepackage{epsfig}
\usepackage{graphicx}
\usepackage{amsmath}
\usepackage{amssymb}
\usepackage{multirow}
\usepackage{xcolor}
\usepackage{color}
\usepackage{subeqnarray}
\usepackage{array}
\usepackage{booktabs}
\usepackage{colortbl}
\usepackage{cases}
\usepackage{bm}
\usepackage{stfloats}
\usepackage{subfigure}
\usepackage[misc]{ifsym}
\usepackage{bbding}
\usepackage[pagebackref=true,breaklinks=true,colorlinks,bookmarks=false]{hyperref}
\graphicspath{{fig/}}

\begin{document}

\title{Pseudo-ISP: Learning Pseudo In-camera Signal Processing Pipeline from A {Color} Image Denoiser}

\author{
    Yue Cao$^{1}$\ \ \ Xiaohe Wu$^{1}$\ \ \ Shuran Qi$^{1}$\ \ \ Xiao Liu$^{3}$\ \ \ Zhongqin Wu$^{3}$\ \ \ Wangmeng Zuo$^{1,2 \text{ \Envelope}}$\ \ \
	\\
	\\
    $^{1}$Harbin Institute of Technology, China \quad  $^{2}$Peng Cheng Lab, China\\
    $^{3}$Tomorrow Advancing Life\\
    \footnotesize{cscaoyue@gmail.com, csxhwu@gmail.com,  srqi@hit.edu.cn, ender.liux@gmail.com, 30388514@qq.com, wmzuo@hit.edu.cn}
}


\maketitle


\begin{abstract}

The success of deep denoisers on real-world {color} photographs usually relies on the modeling of sensor noise and in-camera signal processing (ISP) pipeline.
Performance drop will inevitably happen when the sensor and ISP pipeline of test images are different from those for training the deep denoisers (\ie, noise discrepancy).
In this paper, we present an unpaired learning scheme to adapt a {color} image denoiser for handling test images with noise discrepancy.
We consider a practical training setting, \ie, a pre-trained denoiser, a set of test noisy images, and an unpaired set of clean images.
To begin with, the pre-trained denoiser is used to generate the pseudo clean images for the test images.
Pseudo-ISP is then suggested to jointly learn the pseudo ISP pipeline and signal-dependent {rawRGB} noise model using the pairs of test and pseudo clean images.
We further apply the learned pseudo ISP and {rawRGB} noise model to clean color images to synthesize realistic noisy images for denoiser adaption.
Pseudo-ISP is effective in synthesizing realistic noisy sRGB images, and improved denoising performance can be achieved by alternating between Pseudo-ISP training and denoiser adaption.
%
%
Experiments show that our Pseudo-ISP not only can boost simple Gaussian blurring-based denoiser to achieve competitive performance against CBDNet, but also is effective in improving state-of-the-art deep denoisers, \eg, CBDNet and RIDNet.
The source code and pre-trained model are available at  \emph{\url{https://github.com/happycaoyue/Pseudo-ISP}}.
\end{abstract}
\begin{figure}[htbp]
\scriptsize{
\begin{center}
\vspace{-0ex}
\begin{overpic}[width=0.47\textwidth,scale=0.5]{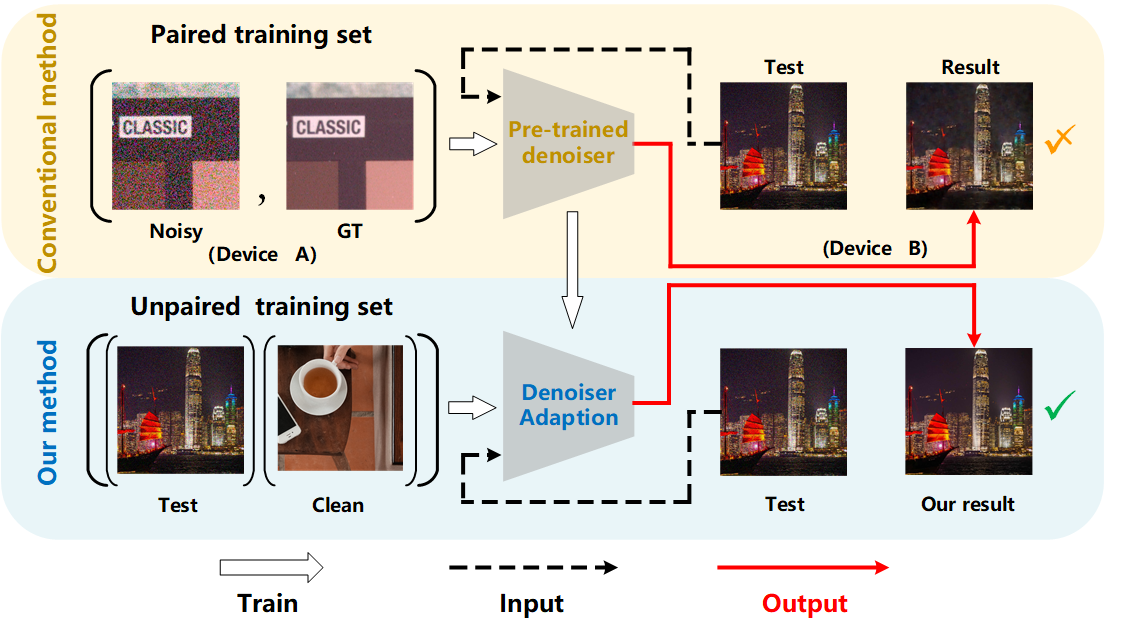} 
\end{overpic}
\end{center}}
\vspace{-4.5ex}
\caption{\small
{Illustration of noise discrepancy and our solution. Pre-trained denoiser for Device $A$ performs inferior on noisy images from Device $B$. Our method utilizes a set of test noisy images and an unpaired set of clean images to adapt the pre-trained denoiser.}
}
\vspace*{-5mm}
\label{fig:scheme}
\end{figure}
\vspace*{-5mm}
\section{Introduction}
\vspace*{-2mm}
Recent years have witnessed the great success of deep convolutional neural networks (CNNs) in additive white Gaussian noise (AWGN) removal~\cite{DnCNN, MWCNN, RIDNet, MemNet}. Subsequently, numerous methods have been developed for handling more sophisticated types of image noise~\cite{islam2018mixed,remez2018class}.
However, these approaches usually are overfitted to the specific noise distribution used in training, and degrade dramatically when applied to real-world photographs.
Actually, real noise is sophisticated and the camera image signal processing (ISP) pipeline further increases its complexity.
As a remedy, existing deep denoisers for handling real-world noisy images usually are trained either by exploiting realistic noise model~\cite{CBDNet,UPI} to synthesize noisy images or by acquiring real paired noisy and noise-free images~\cite{SIDD,Nam,DND}.
However, noise characteristics may vary greatly for different camera sensors and ISO settings, and the ISP pipeline is also device-dependent.
Performance drop will inevitably happen and only limited success will be achieved when applied a deep denoiser to the devices with different sensors and ISP pipelines, \ie, noise discrepancy (see Fig.~\ref{fig:scheme}).
One direct solution is to finetune the pre-trained denoiser by collecting extra noisy-clean image pairs similar to the testing scenario, but it is expensive and unfriendly to practitioners.
Instead, Zamir \etal~\cite{CycleISP} presented a learning-based device-agnostic ISP.
However, it requires a large amount of both paired noisy-clean sRGB images and paired {rawRGB}-{sRGB} data, and cannot generalize well to unseen devices.
To tackle the noise discrepancy issue, this paper presents an unpaired learning scheme to adapt a {color} image denoiser for handling test images with noise discrepancy.
We consider a practical training setting, \ie, a pre-trained denoiser, a set of test noisy images, and an unpaired set of clean images.
We argue that such setting is accessible in practice.
For example, there are several deep denoisers~\cite{CBDNet,RIDNet} that exhibit reasonable denoising and generalization ability on real-world photographs, and it is practically feasible to collect unpaired noisy and clean images.
In general, our unpaired learning scheme alternates between two modules, \ie, learning noise modeling and denoiser adaption.
On the one hand, we exploit the  pre-trained denoiser to generate the pseudo clean images for test images, which are then leveraged for learning noise modeling.
On the other hand, we also apply the learned noise model on clean images to synthesize realistic noisy images for denoiser adaption.
While denoiser adaption can be readily delivered given noisy and clean image pairs, it remains a challenging issue to learn sRGB noise modeling given the test noisy and pseudo clean images.
%
%
While the {rawRGB} image noise can be assumed to be signal-dependent and spatially independent, it is difficult to convert an sRGB image to the rawRGB space due to the unknown ISP pipeline.
To tackle this issue, we present a Pseudo-ISP model involving three subnets, \ie, sRGB2Raw, Raw2sRGB, and noise estimation.
In particular, sRGB2Raw is used to imitate inverse ISP for making the noise to be signal-dependent and spatially independent in the pseudo rawRGB space.
Then, we stack $1\times1$ convolutional layers to form the noise estimation subnet for noise modeling in the pseudo {rawRGB} space.
Finally, Raw2sRGB is deployed to imitate ISP for converting the pseudo {rawRGB} image to color image.
%
%
The learned pseudo ISP and rawRGB noise model can be used to generate realistic noisy sRGB images to benefit denoiser adaption.
%
%
%
%
%

Experiments on five datasets of real-world noisy photographs show that our method performs favorably in terms of quantitative and qualitative results.
Our Pseudo-ISP can not only boost Gaussian blurring-based denoiser to achieve competitive performance, but also improve state-of-the-art deep denoisers, \eg, CBDNet~\cite{CBDNet}, RIDNet~\cite{RIDNet} and PT-MWRN~\cite{cao2020progressive}.
The main contribution of this work includes:
\vspace*{-3mm}
\begin {itemize}
   \vspace*{-3mm}
   \item Equipped with a set of test noisy images and an unpaired set of clean images, an unpaired learning scheme is presented to adapt a {color} image denoiser for handling test images with noise discrepancy.
   \vspace*{-3mm}
   \item Given test noisy and pseudo clean image pairs, a Pseudo-ISP model is suggested to jointly learn the pseudo ISP pipeline and pseudo {rawRGB} noise model for noise modeling of real-world sRGB images.
   \vspace*{-3mm}
   \item Experiments show that our approach can be incorporated with either weak (\eg, Gaussian blurring) or state-of-the-art (\eg, RIDNet) denoisers to boost denoising performance on test noisy images.
 \end {itemize}

\vspace*{-4mm}
\section{Related Work}
\label{sec:related_work}
%
\vspace*{-1mm}
\subsection{Denoising of Real-world Photographs}
\vspace*{-2mm}
In the recent past, great progress of CNN denoisers have been made in AWGN noise removal~\cite{DnCNN, FFDNet}.
Advanced methods have been intensively studied by improving network architectures~\cite{ronneberger2015u, He2016Deep, yu2019deep, park2019densely}
and introducing efficient modules such as dilated convolution~\cite{zhang2017learning}, attention mechanism~\cite{RIDNet,kim2019grdn,Zamir2020MIRNet} and wavelet transform~\cite{MWCNN}.
However, such data-driven approaches are prone to be overfitted to the synthetic training data from specific noise model. 
For handling real-world noisy images, one usual solution is to leverage large-scale paired images for supervised learning.
But it remains a challenging issue to collect paired images.
%
%
Several approaches have been suggested to capture nearly clean images by averaging a burst of noisy images~\cite{SIDD,Nam} or post-processing the long exposure image~\cite{DND}.
However, such data acquisition methods are cost-expensive and time consuming.
And the acquired noise-free images may suffer from over-smoothing issue due to the averaging effect.
Efforts have been made on synthesizing realistic noisy images~\cite{CBDNet, UPI, NoiseFlow, CycleISP}.
CBDNet~\cite{CBDNet} presents a realistic noise model including heterogenous Gaussian and ISP pipeline.
%
UPI~\cite{UPI} further details the ISP pipeline and presents a systematic approach for modeling these key components.
These physical camera ISP based methods overly depend on target device, and the trained denoisers may perform limited when deployed to the device with different {imaging sensors} and ISP pipelines.
Instead of explicit ISP modeling, Zamir \etal~\cite{CycleISP} present a learning-based device-agnostic ISP.
However, it requires plenty of paired noisy-clean sRGB images and {sRGB}-{rawRGB} data, and may not be extended to unseen devices well.
%

\vspace*{-2mm}
\subsection{Noise Modeling of Real-world Photographs}
%
%
\vspace*{-1mm}
%
Though many attempts have been made on conventional image noise~\cite{DnCNN,islam2018mixed,remez2018class}, they generally are limited in handling real-world noise.
In practice, noise of real-world photographs is complicated, and is affected by both camera sensors and ISP pipeline.
Sensor noise stems from various sources.
Considering the primary photon sensing and stationary disturbances, Gaussian-Poisson and heteroscedastic Gaussian are widely employed to characterize the {rawRGB} noise~\cite{CBDNet,UPI,CycleISP}.
%
Most recently, more sophisticated sensor noise model are explored.
Wang~\emph{et al.}~\cite{wang2020Practical} propose an ISO-dependent noise model to simulate the high-sensitivity noise in real-world {sRGB} images.
Wei~\emph{et al.}~\cite{wei2020physical} present a physics-based noise formation model derived from electronic imaging pipeline in a fine-grained manner.
Explicit noise modeling may be overfitted to specific noise and cannot fully characterize the complexity of real-world image noise.
%
Recent studies~\cite{NoiseFlow} show that it is feasible to learn noise model benefitting from the modeling capability of CNNs.
%
%
Besides, GAN-based generative model provides an alternative to characterize noise distribution~\cite{chen2018GCBD, kim2019grdn, chang2020noisegan}.
%
%
However, existing supervised noise models generally require both paired noisy-clean images and paired {rawRGB}-{sRGB} data, limiting their practicality.
%
%
%

\vspace*{-1mm}
\subsection{Self-supervised Image Denoising}
\vspace*{-1mm}
Self-supervised denoisers have drawn much recent attention.
Zhussip~\emph{et al.}~\cite{SURE} adopt the Steins unbiased risk estimator (SURE) to learn CNN denoisers from pairs of noisy images,
while it is limited to AWGN noise removal and noise level should be given as prior.
Lehtinen~\emph{et al.}~\cite{N2N} suggest a Noise2Noise (N2N) model but require paired noisy image.
Recently, blind-spot network (BSN) based denoisers~\cite{N2S,N2V} provide an interesting solution by using only noisy images in training, but suffer from the training inefficiency issue.
%
%
Moreover, they fail to exploit the noisy pixel value at the same position in input, giving rise to degraded performance.
Subsequently, masked convolution~\cite{HQS} and probabilistic inference~\cite{HQS, PN2V} are further introduced for improving denoising performance.
%
%
%
DBSN~\cite{DBSN} extends the noise model to be pixel-independent and signal-dependent, further presents an unpaired learning framework of deep denoising networks.
However, the assumed noise model ignores the influence of ISP pipeline, and only achieves limited success on handling real-world noisy photographs.
%
%

\vspace*{-3mm}
\section{Proposed Method}
\label{sec:method}
\vspace*{-1mm}
We first explain our motivation. Then, a brief description is presented to adapt pre-trained denoiser for handling noise discrepancy. Finally, we specifically describe Pseudo-ISP for noise modeling and realistic noisy image synthesis.
%

\vspace*{-1mm}
\subsection{Motivation}\label{sec:motivation}
%
%
%
\vspace*{-1mm}
To adapt a pre-trained denoiser to test images with noise discrepancy, we present an unpaired learning scheme by incorporating a pre-trained denoiser with an unpaired set of clean and test noisy images.
%
%
%
We argue that such problem setting is practically feasible.
First, it is feasible to collect a set of test noisy images and another set of high-quality clean images in practice.
Second, with the progress of image denoising, several existing denoisers (\eg,~\cite{CBDNet,RIDNet}) have exhibited reasonable denoising and generalization ability to test images with different sensors and ISP pipelines.
Moreover, in comparison to unpaired learning with only clean and noisy images, our method can further make use of the pre-trained denoiser to generate pseudo clean images, thereby being beneficial to the learning of noise model.

%
%
%
%

%

With such problem setting, there remains several challenging issues.
First, the sRGB image noise is sophisticated and spatially correlated, making it difficult to be modeled from paired noisy and pseudo clean images.
Nonetheless, real-world sRGB image is obtained by passing rawRGB image through ISP pipeline, and the rawRGB image noise is usually assumed to be spatially independent.
Thus, we suggest to exploit a Pseudo-ISP model to convert an sRGB image to pseudo rawRGB space and vice versa.
And $1\times1$ CNN is deployed for learning pixel-wise noise model in the pseudo rawRGB space.
%
%
As explained in Sec.~\ref{sec:noisy_image_generation},
when the necessary assumptions are satisfied, Pseudo-ISP can guarantee to generate realistic noisy sRGB images.
Second, proper adaption is also required to enhance pre-trained denoiser for improving denoising performance on test images.

%
%
%
%

\begin{figure}[!t]
\scriptsize{
\begin{center}
\vspace{-0ex}
\begin{overpic}[width=0.48\textwidth,scale=0.5]{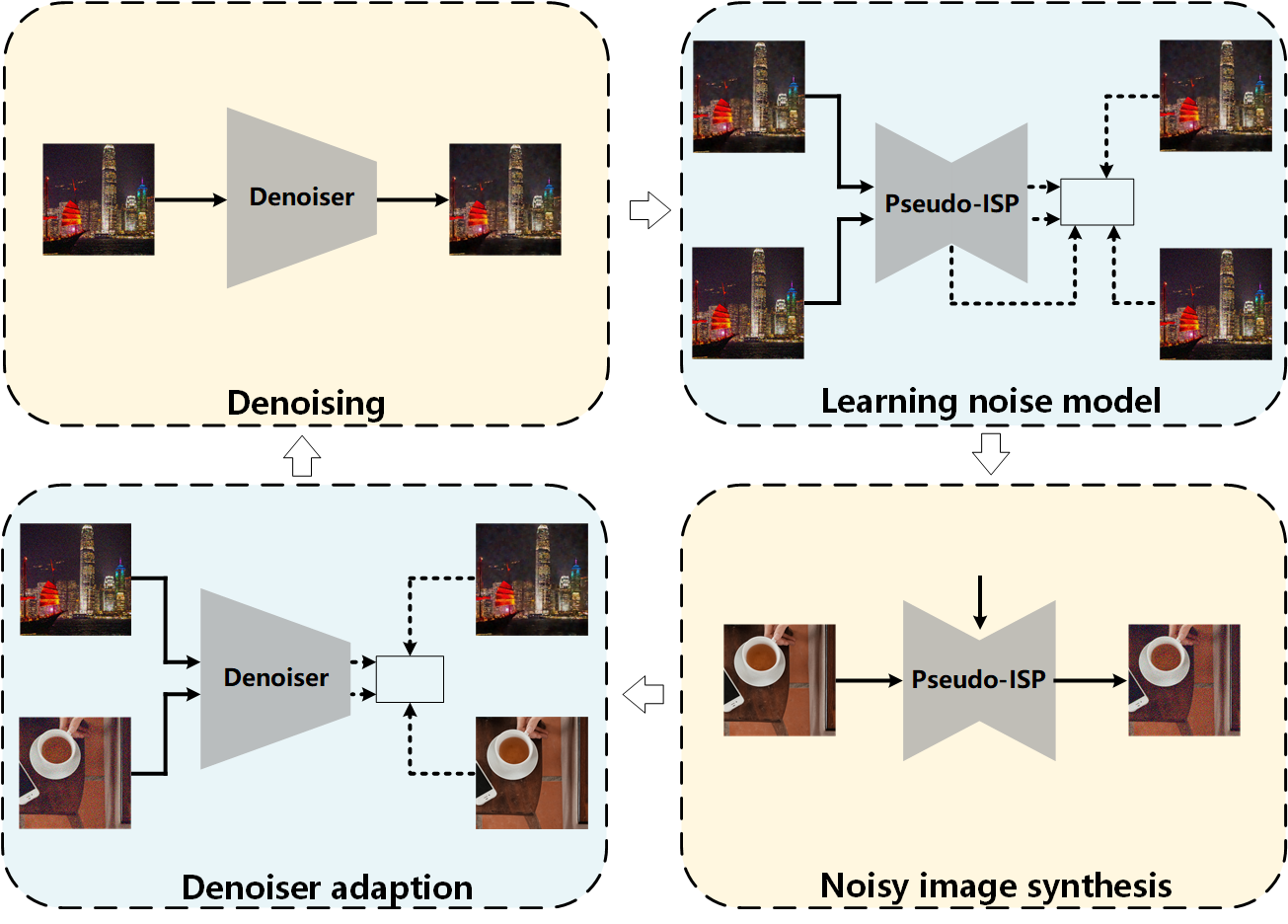} 
\put(6.2,49.2){\color{black}{\tiny $\mathbf{Y}$}}
\put(5.2,47.2){\color{black}{\tiny (test)}}
\put(38,49.2){\color{black}{\tiny $\mathbf{\hat{X}}$}}
\put(33,47.2){\color{black}{\tiny (pseudo clean)}}

\put(57.2,57.3){\color{black}{\tiny $\mathbf{\hat{X}}$}}
\put(52.7,55.3){\color{black}{\tiny (pseudo clean)}}
\put(93,57.3){\color{black}{\tiny $\mathbf{\hat{X}}$}}
\put(88,55.3){\color{black}{\tiny (pseudo clean)}}
\put(57.2,41.7){\color{black}{\tiny $\mathbf{Y}$}}
\put(56.2,39.6){\color{black}{\tiny (test)}}
\put(93,41.7){\color{black}{\tiny $\mathbf{Y}$}}
\put(92.1,39.6){\color{black}{\tiny (test)}}
\put(83.2,55){\color{black}{$\mathcal{L}$}}
\put(85.3,54.7){\scalebox{0.75}{\color{black}{\tiny $\mathcal{P}$}}}

\put(4.7,19.7){\color{black}{\tiny $\mathbf{Y}$}}
\put(3.7,17.7){\color{black}{\tiny (test)}}
\put(40,19.7){\color{black}{\tiny $\mathbf{\hat{X}}$}}
\put(34.8,17.7){\color{black}{\tiny (pseudo clean)}}
\put(4.7,4.7){\color{black}{\tiny $\mathbf{\hat{Y}}$}}
\put(1.6,2.7){\color{black}{\tiny (pseudo noisy)}}
\put(40,4.7){\color{black}{\tiny $\mathbf{X}$}}
\put(38.5,2.7){\color{black}{\tiny (clean)}}
\put(29.2,18){\color{black}{ $\mathcal{L}$}}
\put(31.9,17.5){\scalebox{0.75}{\color{black}{\tiny $\mathcal{D}$}}}

\put(59,11.9){\color{black}{\tiny $\mathbf{X}$}}
\put(57.5,9.9){\color{black}{\tiny (clean)}}
\put(91,11.9){\color{black}{\tiny $\mathbf{\hat{Y}}$}}
\put(86,9.9){\color{black}{\tiny (pseudo noisy)}}
\put(69,27.7){\color{black}{\tiny $\mathbf{n}_{0} \sim \mathcal{N}(0,1)$}}
\end{overpic}
\caption{\small
{Illustration of our unpaired learning scheme, which iterates with four steps.
First, the denoiser is used to obtain pseudo clean images of test noisy images.
Then, Pseudo-ISP is deployed to learn noise model in the pseudo rawRGB space, which is further used to synthesize realistic noisy images.
Finally, the denoiser is finetuned for adaption using both pseudo and synthetic paired data.}}
\label{fig:unpaired_learning}
\vspace{-9mm}
\end{center}}
\end{figure}
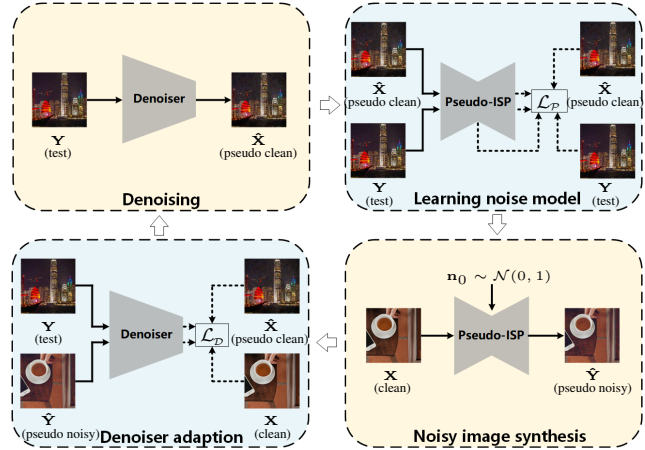
%
%
\vspace*{-2mm}
\subsection{Unpaired Learning Scheme}
\label{sec:unpaired_learning}
\vspace*{-2mm}
We present an unpaired learning scheme by using a pre-trained denoiser, a set of test noisy images $\mathcal{Y}$, and an unpaired set of clean images $\mathcal{X}$.
Denote by $\mathbf{Y}$ a test noisy image from $\mathcal{Y}$, and $\mathbf{X}$ a clean image from $\mathcal{X}$.
Notably, the real noisy observation of $\mathbf{X}$ is unavailable, and so does the noise-free image of $\mathbf{Y}$.

%
%
%

As illustrated in Fig.~\ref{fig:unpaired_learning}, the unpaired learning scheme is achieved by iterating between four steps.
To begin with, we apply a pre-trained denoiser on the test noisy image $\mathbf{Y}$, and obtain the corresponding pseudo clean image $\mathbf{\hat{X}}$.
It allows us to build a set of paired noisy and pseudo clean images, denoted by $\{(\mathbf{\hat{X}}, \mathbf{Y}) | \mathbf{Y} \in \mathcal{Y}\}$.
By leveraging the pseudo paired images, Pseudo-ISP  jointly learns a pseudo ISP pipeline and signal-dependent rawRGB noise model in the pseudo rawRGB space (see Sec.~\ref{sec:pseudoisp}).
Then, given a clean image $\mathbf{X} \in \mathcal{X}$, Pseudo-ISP can be utilized to produce a synthetic noisy image $\mathbf{\hat{Y}}$ (see Sec.~\ref{sec:noisy_image_generation}).
%
Consequently, we build the second set of paired images $\{(\mathbf{{X}}, \mathbf{\hat{Y}}) | \mathbf{X} \in \mathcal{X}\}$.
To adapt the pre-trained {color} denoiser for handling test images, we make use of the above two paired sets to finetune
the denoiser by minimizing the following loss:
\begin{equation}
\small
\setlength{\abovedisplayskip}{4pt}
\setlength{\belowdisplayskip}{4pt}
\label{eq:loss_retraining}
\begin{split}
\mathcal{L}_{D} =
\Big\| {\mathbf{O}}_{\mathbf{\hat{Y}}} - {\mathbf{X}} \Big\|^{2} +
\Big\| {\mathbf{O}}_{\mathbf{{Y}}} - {\mathbf{\hat{X}}} \Big\|^{2} \;,
\end{split}
\end{equation}
where ${\mathbf{O}}_{\mathbf{\hat{Y}}}$ denotes the output of {color} denoiser for the synthetic noisy input $\mathbf{\hat{Y}}$, and ${\mathbf{O}}_{\mathbf{{Y}}}$ is the output of {color} denoiser for test noisy image $\mathbf{Y}$.
\begin{figure*}[!t]
\scriptsize{
\begin{center}
\vspace{-0ex}
\begin{overpic}[width=0.95\textwidth,scale=0.5]{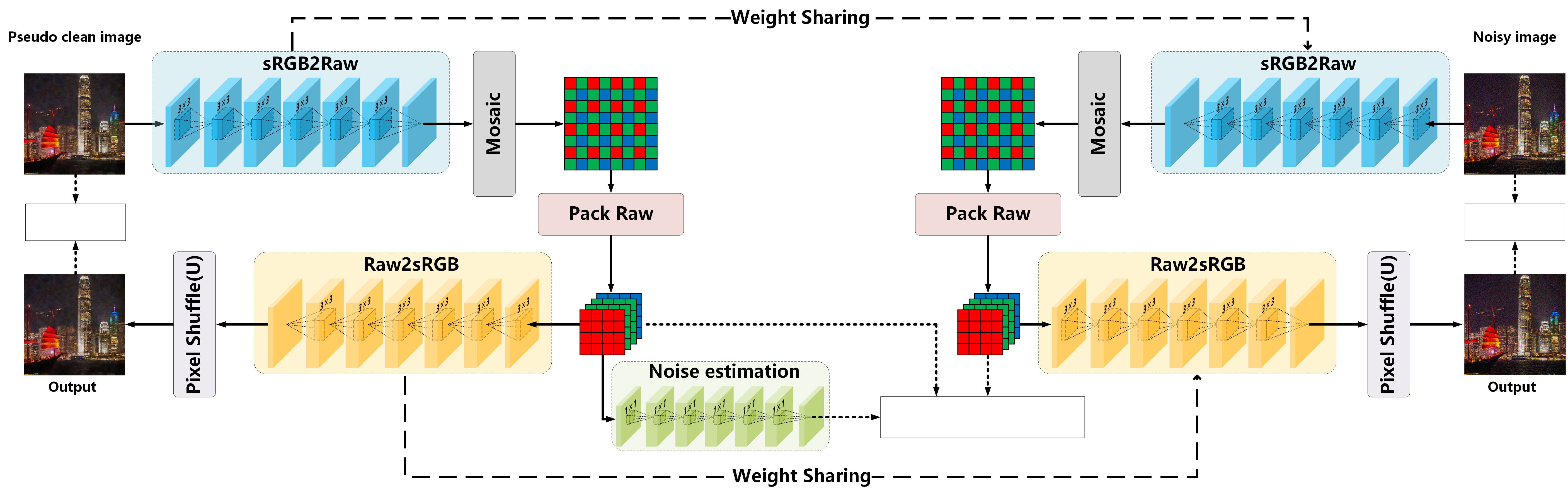} 
\put(2.1,27.3){\color{black}{\tiny $\mathbf{\hat{X}} \!\in\! \mathbb{R}$}}
\put(5.3,28){\scalebox{0.55}{\color{black}{\tiny ${H \!\!\times \!\!W \!\!\times \!\!3}$}}}
\put(10.5,19.8){\color{black}{\tiny 3}}
\put(12.6,19.8){\color{black}{\tiny 128}}
\put(15.1,19.8){\color{black}{\tiny 128}}
\put(17.6,19.8){\color{black}{\tiny 128}}
\put(20.1,19.8){\color{black}{\tiny 128}}
\put(22.7,19.8){\color{black}{\tiny 128}}
\put(25.7,19.8){\color{black}{\tiny 3}}
\put(26,27.4){\color{black}{\tiny $\mathbf{\hat{X}}_{dem}$}}
\put(36,27.4){\color{black}{\tiny $\mathbf{\hat{X}}_{raw}\!\!\in\! \mathbb{R}$}}
\put(41.7,28){\scalebox{0.55}{\color{black}{\tiny ${H\!\!\times \!\!W\!\!\times \!\!1}$}}}

\put(39.3,14.3){\color{black}{\tiny $\mathbf{\hat{X}}_{pack}\!\!\in\! \mathbb{R}$}}
\put(45.4,15){\scalebox{0.55}{\color{black}{\tiny ${  \frac{H}{2} \!\!\times\!\! \frac{W}{2}\!\! \times\!\! 4}$}}}

\put(32.1,6.9){\color{black}{\tiny 4}}
\put(29.2,6.9){\color{black}{\tiny 128}}
\put(26.7,6.9){\color{black}{\tiny 128}}
\put(24.1,6.9){\color{black}{\tiny 128}}
\put(21.6,6.9){\color{black}{\tiny 128}}
\put(19.2,6.9){\color{black}{\tiny 128}}
\put(16.9,6.9){\color{black}{\tiny 12}}
\put(2.1,5.1){\color{black}{\tiny $\mathbf{\hat{X}}^{*}\!\!\in\! \mathbb{R}$}}
\put(5.9,5.8){\scalebox{0.55}{\color{black}{\tiny ${H \!\!\times \!\!W \!\!\times \!\!3}$}}}

\put(1.7,17.2){\color{black}{\tiny $\left \| \hat{\mathbf{X}}^{*} \!\!-\!\! \hat{\mathbf{X}}  \right \|^{2}$}}

\put(93.7,27.3){\color{black}{\tiny $\mathbf{Y}\!\in\! \mathbb{R}$}}
\put(96.9,28){\scalebox{0.55}{\color{black}{\tiny ${H \!\!\times \!\!W \!\!\times \!\!3}$}}}

\put(89.4,19.8){\color{black}{\tiny 3}}
\put(86.4,19.8){\color{black}{\tiny 128}}
\put(83.9,19.8){\color{black}{\tiny 128}}
\put(81.3,19.8){\color{black}{\tiny 128}}
\put(78.8,19.8){\color{black}{\tiny 128}}
\put(76.3,19.8){\color{black}{\tiny 128}}
\put(74.2,19.8){\color{black}{\tiny 3}}
\put(72.1,27.4){\color{black}{\tiny $\mathbf{Y}_{dem}$}}
\put(60,27.4){\color{black}{\tiny $\mathbf{Y}_{raw}\!\!\in\! \mathbb{R}$}}
\put(65.7,28){\scalebox{0.55}{\color{black}{\tiny ${H\!\!\times \!\!W\!\!\times \!\!1}$}}}

\put(53.2,14.3){\color{black}{\tiny $\mathbf{Y}_{pack}\!\!\in\! \mathbb{R}$}}
\put(59.3,15){\scalebox{0.55}{\color{black}{\tiny ${  \frac{H}{2} \!\!\times\!\! \frac{W}{2}\!\! \times\!\! 4}$}}}
\put(67,6.9){\color{black}{\tiny 4}}
\put(69.1,6.9){\color{black}{\tiny 128}}
\put(71.6,6.9){\color{black}{\tiny 128}}
\put(74.1,6.9){\color{black}{\tiny 128}}
\put(76.8,6.9){\color{black}{\tiny 128}}
\put(79.2,6.9){\color{black}{\tiny 128}}
\put(81.9,6.9){\color{black}{\tiny 12}}

\put(93.7,5.1){\color{black}{\tiny $\mathbf{Y}^{*}\!\!\in\! \mathbb{R}$}}
\put(97.5,5.8){\scalebox{0.55}{\color{black}{\tiny ${H \!\!\times \!\!W \!\!\times \!\!3}$}}}

\put(93.5,17.2){\color{black}{\tiny $\Big \| \mathbf{Y}^{*} \!\!-\!\! \mathbf{Y}  \Big \|^{2}$}}

\put(39.3,2){\color{black}{\tiny 4}}
\put(40.7,2){\color{black}{\tiny 128}}
\put(42.7,2){\color{black}{\tiny 128}}
\put(44.6,2){\color{black}{\tiny 128}}
\put(46.4,2){\color{black}{\tiny 128}}
\put(48.3,2){\color{black}{\tiny 128}}
\put(50.8,2){\color{black}{\tiny 4}}
\put(51.4,7.3){\color{black}{ \tiny $\hat{\boldsymbol{\sigma}} \!\in\! \mathbb{R}$}}
\put(54.6,8){\scalebox{0.55}{\color{black}{\tiny ${  \frac{H}{2} \!\!\times\!\! \frac{W}{2} \!\!\times\!\! 4}$}}}

\put(56.3,4.8){\scalebox{0.86}{\color{black}{\tiny $\Big \| \hat{\boldsymbol{\sigma}} \!\!-\!\! \sqrt{\frac{\pi}{2}} | \mathbf{Y}_{\text {pack}} \!\!-\!\! \mathbf{\hat{X}}_{\text {pack}} |\Big \| ^{2}$}}}

\end{overpic}

\caption{\small
{Our Pseudo-ISP learns the pseudo forward and reverse ISP jointly with a pseudo rawRGB noise model. It is composed of three subnets: sRGB2Raw, Raw2sRGB and noise estimation. sRGB2Raw converts an {sRGB} image to the pseudo rawRGB space, in which the noise estimation model is deployed to learn signal-dependent rawRGB noise model.}}
\label{fig:PesudoISP}
\vspace{-8mm}
\end{center}}
\end{figure*}
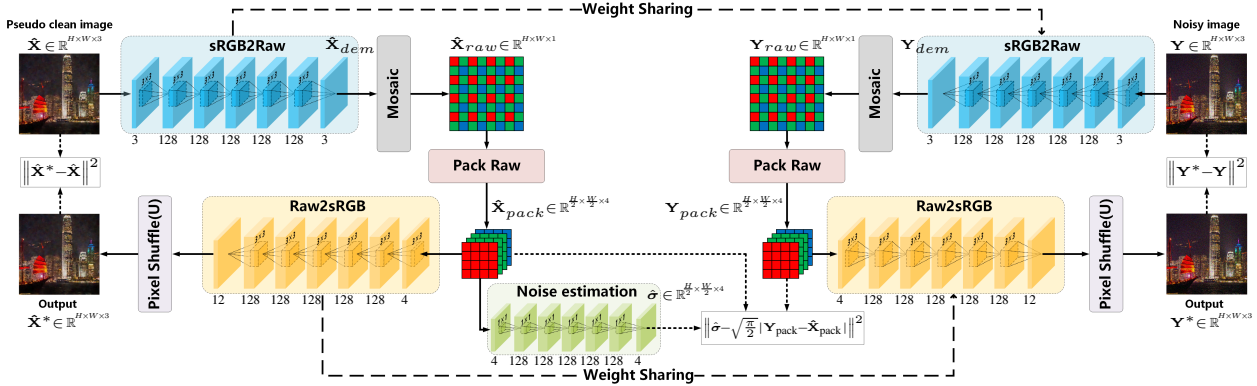


It is noteworthy that the above steps can be iterated for several times to achieve better denoising results.
On the one hand, the adapted denoiser facilitates better pseudo clean images, making that better noise model can be achieved by Pseudo-ISP.
On the other hand, the improved Pseudo-ISP generates more realistic noisy images, then benefiting the subsequent denoiser adaption.
In such an alternating manner, both Pseudo-ISP and denoiser adaption can be improved, thereby resulting in better denoising performance.

\subsection{Learning Pseudo-ISP for Noise Modeling}
\label{sec:pseudoisp}
\vspace*{-1mm}
%
For learning noise model from the pseudo paired images, it is infeasible to learn a direct mapping to predict $\mathbf{Y}$ from $\hat{\mathbf{X}}$ due to the intrinsic randomness of image noise.
Moreover, the sRGB noise is spatially correlated, and thus cannot be characterized with a noise level function (NLF) as in~\cite{DBSN}.
Fortunately, the sRGB image noise is mainly affected by the camera sensors and ISP pipeline.
Albeit the rawRGB noise model is unknown, it generally can be assumed to be signal-dependent and spatially independent~\cite{NoiseFlow}.
The ISP pipeline can be treated as a deterministic mapping from rawRGB image to {sRGB} image.
Here we further assume that the ISP pipeline is reversible, \ie, the rawRGB image can be recovered from {sRGB} image.
Taking these into account, we constitute our Pseudo-ISP involving three subnets, \ie, sRGB2Raw, Raw2sRGB and noise estimation (see Fig.~\ref{fig:PesudoISP}).

%
%
%

%
Denote by the ground-truth clean rawRGB image $\mathbf{X}^{GT}_{raw}$, since the noise is assumed to be signal-dependent and spatially independent, the ground-truth noisy rawRGB image $\mathbf{Y}^{GT}_{raw}$ at pixel $i$ can then be written as:
\vspace*{-0.5mm}
\begin{equation}
\small
\setlength{\abovedisplayskip}{5pt}
\setlength{\belowdisplayskip}{5pt}
\label{eq:noisy_img}
\mathbf{Y}^{GT}_{raw}[i] = \mathbf{X}^{GT}_{raw}[i] + {\mathbf{n}}[i],
\end{equation}
where $\mathbf{n}[i] \!\!\sim\!\! \mathcal{N}(0, \bm{\sigma}^{2}[i])$ denotes the rawRGB noise at pixel $i$ with variance $\bm{\sigma}^{2}[i]$.
Moreover, it is noted that the noise variance at each pixel is determined only by its corresponding noise-free pixel value. That is, at pixel $i$, we have:
\begin{equation}
\small
\setlength{\abovedisplayskip}{5pt}
\setlength{\belowdisplayskip}{5pt}
\label{eq:raw_noise}
\bm{\sigma}^{2}[i] = g(\mathbf{X}^{GT}_{raw}[i])
\end{equation}
Accordingly, $g(\mathbf{X}^{GT}_{raw})$ can be regarded as the NLF. 
%
%
Interestingly, $g(\mathbf{X}^{GT}_{raw})$ can be represented as a noise estimation network by stacking $1 \times 1$ {group} convolutional layers with group number of 4, and can be learned from a pair of real noisy and clean rawRGB images using the following loss,
\begin{equation}
\small
\setlength{\abovedisplayskip}{5pt}
\setlength{\belowdisplayskip}{5pt}
\label{eq:loss_noise}
\mathcal{L}_n = \Big \| \sqrt{g(\mathbf{X}^{GT}_{raw})} - \sqrt{\frac{\pi}{2}} | \mathbf{X}^{GT}_{raw} - \mathbf{Y}^{GT}_{raw} | \Big \|^2.
\end{equation}
%
%
{The term $| \mathbf{X}^{GT}_{raw} - \mathbf{Y}^{GT}_{raw} |$ denotes an entry-wise absolute value operation  which does not change image size.}
Moreover, $| \mathbf{X}^{GT}_{raw}[i] - \mathbf{Y}^{GT}_{raw}[i] |$ obeys folded normal distribution \cite{leone1961folded}.
Thus, the corresponding mean of $| \mathbf{X}^{GT}_{raw}[i] - \mathbf{Y}^{GT}_{raw}[i] |$ is $\sqrt{\frac{2}{\pi}} \bm{\sigma}[i]$,
and we then utilize $\sqrt{\frac{\pi}{2}}| \mathbf{X}^{GT}_{raw} - \mathbf{Y}^{GT}_{raw} |$ as supervision for learning noise model.
Motivated by the above analyses, Pseudo-ISP adopts a subnet stacked by six $1\times1$ convolutional layers for noise modeling in the pseudo rawRGB space (see Fig.~\ref{fig:PesudoISP}).
ReLU nonlinearity~\cite{krizhevsky2017imagenet} is deployed for all convolutional layers.
A loss term similar to Eq.~(\ref{eq:loss_noise}) is also adopted for learning pseudo rawRGB noise.
%
%
%

%
%

%
%
%
%
%
%
%
Taking the ISP pipeline into account, we further introduce sRGB2Raw and Raw2sRGB, which collaborate with the noise estimation subnet to form our whole Pseudo-ISP.
In particular, sRGB2Raw and Raw2sRGB are designed for converting an {sRGB} image to the pseudo rawRGB space and vice versa.
Given a paired dataset $\{(\mathbf{\hat{X}}, {\mathbf{Y}}) | {\mathbf{Y}} \in {\mathcal{Y}}\}$, sRGB2Raw imitates the inverse ISP pipeline, and converts an {sRGB} image to the pseudo rawRGB space, in which the noise is assumed to be signal-dependent and spatially independent, and NLF $g(\mathbf{X}_{raw})$ can be learned in a supervised manner.
%
%
%
%
Conversely, Raw2sRGB simulates the ISP pipeline to convert pseudo rawRGB image back to the {sRGB} space.
%
%

%
As shown in Fig.~\ref{fig:PesudoISP}, sRGB2Raw consists of six $3\times3$ convolutional layers followed by ReLU nonlinearity~\cite{krizhevsky2017imagenet}.
Following \cite{CycleISP}, the number of output channels of last layer is set as three to preserve structural information possibly from original image.
It learns the transform $f_{s2raw}(\cdot; \mathbf{W})$, and results in the intermediate outputs, \ie, $({\mathbf{\hat{X}}}_{dem}, {\mathbf{{Y}}}_{dem})$,
\begin{equation}
\small
\setlength{\abovedisplayskip}{5pt}
\setlength{\belowdisplayskip}{5pt}
\label{eq:RGB2dem}
\begin{split}
{\hat{\mathbf{X}}}_{dem} = f_{s2raw}(\hat{\mathbf{X}}; \mathbf{W}),
{\mathbf{{Y}}}_{dem} = f_{s2raw}({\mathbf{Y}}; \mathbf{W})
\end{split}
\end{equation}
Then, Bayer sampling  $f_{CFA}$~\cite{CycleISP} is applied to obtain the mosaicked pseudo rawRGB images, \ie, $({\mathbf{\hat{X}}}_{raw}, {\mathbf{{Y}}}_{raw})$,
\begin{equation}
\small
\setlength{\abovedisplayskip}{5pt}
\setlength{\belowdisplayskip}{5pt}
\label{eq:dem2Raw}
\begin{split}
{\hat{\mathbf{X}}}_{raw} = f_{CFA}({\hat{\mathbf{X}}}_{dem}), {\mathbf{{Y}}}_{raw} = f_{CFA}({\mathbf{{Y}}}_{dem})
\end{split}
\end{equation}
To reduce the computational burden, we pack the $2\times2$ blocks of ${\mathbf{\hat{X}}}_{raw}$ and ${\mathbf{{Y}}}_{raw}$ into four channels, bring forth the packed pseudo rawRGB image pairs $(\hat{\mathbf{X}}_{pack}, {\mathbf{Y}}_{pack})$ with the resolution halved as shown in Fig.~\ref{fig:PesudoISP}.
%

Considering the symmetry of forward and inverse ISP pipeline, Raw2sRGB adopts a similar architecture to sRGB2Raw.
%
It converts the packed pseudo rawRGB images of $(\hat{\mathbf{X}}_{pack}, {\mathbf{Y}}_{pack})$ back to the {sRGB} space,
\begin{equation}
\label{eq:Raw2RGB}
\small
\begin{split}
{\mathbf{\hat{X}}}^{*} \!\!=\!\! PSU(f_{raw2s}(\hat{\mathbf{X}}_{pack}; \mathbf{Q}))\\
{\mathbf{{Y}}}^{*} \!\!=\!\! PSU(f_{raw2s}({\mathbf{Y}}_{pack}; \mathbf{Q}))
\end{split}
\end{equation}
where $f_{raw2s}(\cdot;\mathbf{Q})$ denotes the transform learned by Raw2sRGB with shared weights $\mathbf{Q}$, and $PSU(\cdot)$ represents the {pixel shuffle upsampling} operation~\cite{FFDNet}. $({\mathbf{\hat{X}}}^{*},{\mathbf{{Y}}}^{*})$ are the reconstructed paired images in the {sRGB} space.
To jointly learn the pseudo ISP and pseudo noise model, we design the following loss function,
\begin{equation}
\small
\setlength{\abovedisplayskip}{5pt}
\setlength{\belowdisplayskip}{5pt}
\label{eq:loss_pseudoisp}
\begin{split}
\!\!\!\!\!\!\!
\mathcal{L}_{P} \!\!=\!\!
\Big \| \hat{\mathbf{X}}^{*} \!\!-\!\! \hat{\mathbf{X}}  \Big \|^{2}
\!+\! \Big \| \mathbf{Y}^{*} \!\!-\!\! \mathbf{Y}  \Big \|^{2}
\!+\!\lambda \Big \| \bm{\hat{\sigma}} \!-\!\sqrt{\frac{\pi}{2}} | \mathbf{Y}_{\text {pack}} \!-\! \mathbf{\hat{X}}_{\text {pack}} |\Big \| ^{2}
\end{split}
\end{equation}
where $\lambda$ is a positive constant, and $\bm{\hat{\sigma}}$ donates the output of noise estimation subnet.
%
%
%
%
%

\noindent \textbf{Discussion.} We note that the three terms in Eq.~(\ref{eq:loss_pseudoisp}) collaborate to learn reasonable Pseudo-ISP model.
By assuming that Pseudo-ISP is approximately invertible, we have the first two terms in Eq.~(\ref{eq:loss_pseudoisp}).
The structure of noise estimation network makes it only predict the pixel-wise and signal-dependent component of NLF.
When the noise in pseudo rawRGB space is still spatially correlated, it becomes difficult to estimate the noise level via pixel-wise mapping, and thus the last term will be larger.
Thus, the minimization of the last term is beneficial to learn sRGB2Raw for eliminating the spatial correlation of pseudo rawRGB noise.
%
%
Moreover, both sRGB2Raw and Raw2sRGB subnets involve 6 convolutional layers, indicating that our Pseudo-ISP is able to model complex ISP pipelines~\cite{MIT}.

Pseudo-ISP also differs from CycleISP~\cite{CycleISP} for learning ISP pipeline in a data-driven manner.
%
It depends on paired clean-noisy sRGB images and paired clean rawRGB-{sRGB} images and its performance may degrade when applied to unseen devices.
%
%
In contrast, Pseudo-ISP converts an sRGB image to pseudo rawRGB space and vice versa, in which the pseudo rawRGB noise can be modeled by the signal-dependent and spatially independent noise model.
Most importantly, our Pseudo-ISP only requires unpaired clean-noisy sRGB images, which is practically more feasible.
%
%
%

%
%
%
%
%

%
\begin{figure}[t]
\scriptsize{
\begin{center}
\vspace{-0ex}
\begin{overpic}[width=0.47\textwidth,scale=0.5]{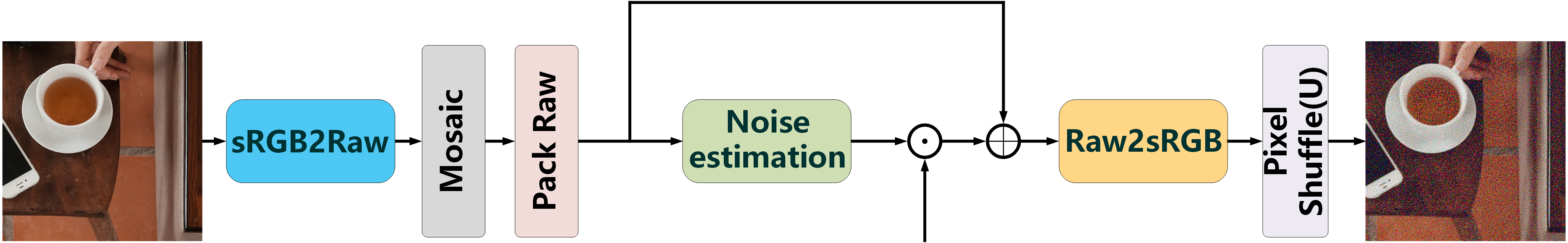} 
\put(5.2,-2.4){\color{black}{\tiny $\mathbf{X}$}}
\put(37.1,3.1){\color{black}{\tiny $\mathbf{X}_{pack}$}}
\put(92.2,-2.4){\color{black}{\tiny $\mathbf{\hat{Y}}$}}
\put(52,-1.5){\color{black}{\tiny $\mathbf{n}_{0} \sim \mathcal{N}(0,1)$}}
\end{overpic}
\vspace{3ex}
\caption{\small
{Synthetic noisy image generation using Pseudo-ISP.}}
\label{fig:gen}
\vspace{-9mm}
\end{center}}
\end{figure}
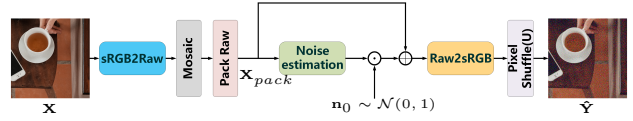
%

\vspace*{-1mm}
\subsection{Synthetic Noisy Image Generation}
\label{sec:noisy_image_generation}
\vspace*{-1mm}
%
%
%
%
%
%
Once the Pseudo-ISP has been trained, we can use it to synthesize realistic noisy image for a given clean image $\mathbf{X}$.
%
%
As shown in Fig.~\ref{fig:gen}, we first use sRGB2Raw to convert a clean observation to the pseudo rawRGB space.
%
%
%
Then, Bayer sampling and packing operation are applied to achieve the packed rawRGB image $\mathbf{X}_{pack}$.
In the pseudo rawRGB space, the estimated noise model $\hat{g}(\mathbf{X}_{pack})$ is used to predict noise standard deviation for $\mathbf{X}_{pack}$.
The noisy pseudo rawRGB image $\mathbf{\hat{Y}}_{pack}$ can then be synthesized by,
\begin{equation}
\small
\setlength{\abovedisplayskip}{5pt}
\setlength{\belowdisplayskip}{5pt}
\label{eq:pack_noise}
\begin{split}
{\mathbf{\hat{Y}}}_{pack} = \mathbf{X}_{pack} + \hat{g}(\mathbf{X}_{pack}) \cdot \mathbf{n}_{0},
\end{split}
\end{equation}
where $\mathbf{n}_{0} \sim \mathcal{N}(0,1)$ is a random noise sampled from normal distribution.
Through Raw2sRGB and pixel shuffle {upsampling}, the synthetic noisy image $\mathbf{\hat{Y}}$ can be attained, and we build the synthetic paired dataset $\{(\mathbf{X}, \mathbf{\hat{Y}}) | \mathbf{X} \in \mathcal{X} \}$.

\noindent \textbf{Discussion.} Our Pseudo-ISP does not require to accurately recover the ground-truth ISP and rawRGB noise model.
Denote by $\mathbf{Y}^{GT}_{raw}$ the ground-truth noisy rawRGB image and $\mathbf{Y}_{raw}$ the pseudo rawRGB image.
When there is an invertible element-wise mapping between $\mathbf{Y}^{GT}_{raw}$ and $\mathbf{Y}_{raw}$, \ie, $\mathbf{Y}_{raw}[i]\!\! = \!\! f\left(\mathbf{Y}^{GT}_{raw}[i]\right)$  and $\mathbf{Y}^{GT}_{raw}[i]\!\! = \!\! f^{\!-\!1}\left(\mathbf{Y}_{raw}[i]\right)$, and the learned noise estimation model is proper,
our Pseudo-ISP is able to approximate the noise models in both rawRGB and sRGB spaces.

To illustrate this point, we assume that a real-world noisy rawRGB image can be written as,
\begin{equation}
\small
\setlength{\abovedisplayskip}{5pt}
\setlength{\belowdisplayskip}{5pt}
\label{eq:noisy_n0}
\mathbf{Y}^{GT}_{raw} = \mathbf{X}^{GT}_{raw} + g(\mathbf{X}^{GT}_{raw}) \cdot \mathbf{n}_{0},
\end{equation}
where $g(\cdot)$ denotes a $1 \times 1$ CNN for deriving ground-truth noise standard deviation.
Consequently, we have,
\begin{equation}
\small
\setlength{\abovedisplayskip}{5pt}
\setlength{\belowdisplayskip}{5pt}
\label{eq:noisy_invf}
\mathbf{Y}_{raw}[i] \!=\! f\left(\mathbf{Y}^{GT}_{raw}[i] \right) \!=\! f\left( \mathbf{X}^{GT}_{raw}[i] \!+\! g(\mathbf{X}^{GT}_{raw}[i]) \!\cdot\! \mathbf{n}_{0}[i] \right) .
\end{equation}
We note that $\mathbf{X}_{raw}[i] = f\left( \mathbf{X}^{GT}_{raw}[i] \right)$.
By approximating the last term with its first order Taylor expansion, the pseudo rawRGB noisy image can be approximated by,
\begin{equation}
\small
\setlength{\abovedisplayskip}{5pt}
\setlength{\belowdisplayskip}{5pt}
\label{eq:noisy_pse}
\begin{split}
\mathbf{Y}_{raw}[i] \approx \mathbf{X}_{raw}[i] + h\left(\mathbf{X}_{raw}[i]\right) \cdot \mathbf{n}_{0}[i],
\end{split}
\end{equation}
where $h(\mathbf{X}_{raw}[i])$ denotes the estimated noise standard deviation in pseudo rawRGB space, and can be obtained by,
\begin{equation}
\small
h(\mathbf{X}_{raw}[i])= f'(f^{-1}(\mathbf{X}_{raw}[i])) \cdot g(f^{-1}(\mathbf{X}_{raw}[i]))
\end{equation}
where $f'$ denotes the first-order derivative of $f$.
Thus, $h(\cdot)$ can also be represented as a $1 \times 1$ CNN and our noise estimation model $\hat{g}(\cdot)$ can serve as an approximation of $h(\cdot)$.
%

%

To sum up, we assume that $(i)$ there is an invertible element-wise mapping for approximating $\mathbf{Y}^{GT}_{raw}$ with $\mathbf{Y}_{raw}$ and vice versa, and $(ii)$ $\hat{g}(\cdot)$ is a good estimation of $h(\cdot)$.
Then, we can use $\hat{g}(\cdot)$ to add noise in pseudo rawRGB space, and utilize $f^{\!-\!1}(\cdot)$ to synthesize realistic noisy image in the ground-truth rawRGB space.
In Sec.~\ref{sec:Assumption_test}, we show that the assumptions empirically hold on, and it is practically feasible to synthesize realistic noisy rawRGB images via Pseudo-ISP.
%
%
Moreover, Pseudo-ISP can guarantee to generate realistic noisy sRGB images.
Denote by $f_{raw2s}^{*}$ and $f_{s2raw}^{*}$ the ground-truth ISP and inverse ISP models.
We then have $f_{s2raw}(\cdot) \approx f(f_{s2raw}^{*}(\cdot))$ and $f_{raw2s}(\cdot) \approx f_{raw2s}^{*}(f^{-1}(\cdot))$.
Consequently, we have,
\begin{equation}
\small
f_{raw2s}(\mathbf{Y}_{raw}) \!\approx\! f_{raw2s}^{*}(f^{-1}(\mathbf{Y}_{raw})) \!\approx\! f_{raw2s}^{*}(\mathbf{Y}_{raw}^{GT}).
\end{equation}
That is, even $f_{s2raw}(\cdot) \neq f_{s2raw}^{*}(\cdot)$, our Pseudo-ISP can also be used to synthesize realistic noisy sRGB images. 


%
%

%
\vspace*{-2mm}
\section{Experiments}
\label{sec:experiments}
\vspace*{-1mm}

\subsection{{Experimental Settings}
}
\label{sec:denoisers_images}
\vspace*{-1mm}
%
%

\noindent \textbf{Pre-trained denoisers.}
Three pre-trained deep models, \ie, CBDNet~\cite{CBDNet}, RIDNet~\cite{RIDNet} and PT-MWRN~\cite{cao2020progressive}, are adopted for evaluation, which are released officially by the authors.
%
%
%
%
Moreover, both traditional and unsupervised denoising methods, \ie, Gaussian blurring, BM3D~\cite{dabov2008image} and DIP~\cite{DIP}, are considered.
For these methods, we re-train MWCNN~\cite{MWCNN} in the first denoiser adaption step and then use it in the alternated training.

\noindent \textbf{Unpaired set of noisy and clean images.}
For real-world noisy images, we use DND~\cite{DND}, SIDD~\cite{SIDD}, SIDDPlus~\cite{ntire2020}, CC15~\cite{Nam} and MIT-IP8~\cite{MIT} as the sets of test noisy images.
DND consists of $50$ pairs of noisy and clean images with high-resolution, while the ground-truth clean data are not publicly available.
Quantitative evaluation can only be performed by an online server\footnote{\scriptsize \url{{https://noise.visinf.tu-darmstadt.de/benchmark/}}}.
SIDD contains three sets for training, validation and testing, respectively.
And the quantitative evaluation on test set can only be performed through an online server\footnote{\scriptsize \url{{https://www.eecs.yorku.ca/~kamel/sidd/benchmark.php}}}.
As an extension of NTIRE2020 challenge on real image denoising, SIDDPlus provides another validation and test sets, in which the noise distribution differs from that in SIDD training.
Due to the {evaluation unavailability} of SIDDPlus test set, we only report the result on its validation set\footnote{\scriptsize \url{{https://bit.ly/siddplus_data}}}.
CC15 is composed of $15$ pairs of noisy and clean patches cropped from Nam~\cite{Nam} with small size $512 \times 512$.
{MIT-IP8 consists of $21$ pairs of noisy and clean iPhone 8 images from~\cite{MIT}.
Following the setting of the other datasets, we randomly crop 35 patches with the size $512 \times 512$ from the original 21 images to constitute MIT-IP8.}
Besides, we take 200 images randomly from DIV2K~\cite{agustsson2017ntire} as the unpaired set of clean images.
%

%
\noindent \textbf{Implementation Details}.
%
We use the Adam optimizer~\cite{adam} for all the models presented in this paper. 
For Gaussian blurring, we fix the size of blur kernel to $5\times5$ and the standard deviation is set as 1.
DIP~\cite{DIP} iterates 3,000 times following its default setting.
%
%
%
We adopt image-specific Pseudo-ISP, \ie, each test noisy image corresponds to one Pseudo-ISP model.
%
%
We randomly crop $12,000 \times 32$ patches with size $60\times60$ to train Pseudo-ISP.
%
%
We use the initial learning rate $10^{-4}$ for $8,000$ iterations and then decrease it to $10^{-5}$ for another $4,000$ iterations.

%
For CBDNet~\cite{CBDNet}, RIDNet~\cite{RIDNet} and PT-MWRN~\cite{cao2020progressive}, both pseudo and synthetic paired images are utilized for denoiser adaption.
The training details, including the batch size and input patch size are the same as their defaults, and the learning rates follow the last epoch of the pre-trained models.
Since Gaussian blurring, BM3D and DIP~\cite{DIP} cannot be trained in a supervised manner, we adopt the randomly initialized MWCNN~\cite{MWCNN} for subsequent denoiser adaption.
\vspace*{-1mm}
\subsection{Assessing Pseudo-ISP Hyper-Parameters}
\label{sec:Ablation_Studies}
\vspace*{-1mm}
%
%
We assess several settings of Pseudo-ISP, including weight sharing, incorporation of pseudo and synthetic paired images, and the times of alternated training.
All the experiments are conducted on DND, and we consider four pre-trained denoisers, \ie, Gaussian blurring, DIP~\cite{DIP}, CBDNet~\cite{CBDNet} and RIDNet~\cite{RIDNet}.

%
\begin{table}[t]
\begin{center}
\caption{PSNR (dB) results obtained using different weight sharing schemes on DND~\cite{DND}. Best results are highlighted.}
\label{tab:image_share_weight}
\vspace{-3mm}
\setlength{\tabcolsep}{6pt}
\scalebox{0.65}{
\begin{tabular}{l c c c c}
\toprule Weight Sharing Scheme & Gaussian Blurring & DIP~\cite{DIP} & CBDNet~\cite{CBDNet} & RIDNet~\cite{RIDNet}\\ \midrule
w/o Weight Sharing & 33.92& 36.05& 38.11& 39.29\\
Patch-specific Sharing & 34.40& 36.20& 38.20& 39.31\\
Image-specific Sharing& \textbf{36.26}& \textbf{37.21}& \textbf{38.59} &\textbf{39.43}\\
Set-specific Sharing& 34.70& 36.43& 38.32& 39.35 \\ \bottomrule

\end{tabular}}
\vspace{-7mm}
\end{center}
\end{table}

\noindent \textbf{Weight Sharing for Learning Pseudo-ISP.}
sRGB2Raw and Raw2sRGB can be used to process either noisy or clean images.
So ablation study is conducted to check whether noisy and clean images could share the weights for sRGB2Raw and Raw2sRGB.
Table~\ref{tab:image_share_weight} lists the PSNR results by patch-specific Pseudo-ISP with and without weight sharing.
For a fair comparison, all the results are obtained by performing denoiser adaption once.
The results indicate that weight sharing benefits denoising performance.
With weight sharing, the number of patches to train sRGB2Raw and Raw2sRGB can be doubled, which explains the improvement on denoising performance.


%

%
%
%
%
%

We further test several other approaches to introduce more patches for training Pseudo-ISP.
Note that each DND image is cropped into 20 patches with the size $512 \times 512$.
So we give the result of image-specific Pseudo-ISP by allowing all the patches from an image share the same weights.
Analogously, set-specific Pseudo-ISP is also provided. 
From Table~\ref{tab:image_share_weight}, more performance gain can be attained by image-specific weight sharing.
Set-specific Pseudo-ISP, however, performs inferior to image-specific one, owing to that the DND images are captured using four different cameras which intrinsically do not share the ISP and noise models.
Thus, image-specific Pseudo-ISP is adopted as the default.


\begin{table}[t]
\begin{center}
\caption{PSNR (dB) results for varying ratios of synthetic paired images per mini-batch (\ie, $r$) for denoiser adaption on DND~\cite{DND}. }
\label{tab:ab r}
\setlength{\tabcolsep}{11pt}
\scalebox{0.68}{
\begin{tabular}{c c c c c}
\toprule
$r$  & Gaussian Blurring & DIP~\cite{DIP} & CBDNet~\cite{CBDNet} & RIDNet~\cite{RIDNet}\\
\midrule
$25\%$&   34.25 &36.30 &    38.47 &39.40 \\
$50\%$&   34.39  &36.71 &    \textbf{38.59}&\textbf{39.43}\\
$75\%$&   34.61 &37.05 &    38.43 & 39.35\\
$100\%$&   \textbf{36.26}&\textbf{37.21} &  38.11 &39.32\\
 \bottomrule
\end{tabular}}
\vspace{-7mm}
\end{center}
\end{table}

\noindent \textbf{Incorporation of Pseudo and Synthetic Paired Images.}
There are two sets of paired images for denoiser adaption, \ie, a pseudo paired set and a synthetic paired set.
Experiments are then conducted by employing different ratios of synthetic noisy images per mini-batch for adapting the pre-trained denoiser.
Table~\ref{tab:ab r} lists the results by setting the ratios to be $25\%$, $50\%$, $75\%$, and $100\%$.
It can be seen that the inclusion of synthetic-noisy set is beneficial to denoising performance.
For traditional and unsupervised methods, the synthetic paired set plays a pivotal role and the best performance is attained by only using synthetic noisy images.
As for pre-trained deep denoisers, the pseudo paired set can serve as a kind of regularization to avoid the overfitting to synthetic paired set and thus is required.
Overall, the best performance can be attained for CBDNet~\cite{CBDNet} and RIDNet~\cite{RIDNet} by using $50\%$ synthetic noisy images.
Thus the ratio setting of $50\%$ is also adopted for other deep denoisers.

%
%
%
%
%
%

%
\noindent \textbf{Times of Alternated Training.}
%
%
As described in Sec.~\ref{sec:unpaired_learning}, the alternating between Pseudo-ISP training and denoiser adaption can be repeated for several times.
%
%
Empirically, increasing the times of alternated training (\ie, $t$) continuously improves denoising performance, and the gains becomes negligible when $t \geq 3$.
Thus, we set $t = 3$.
Please refer to the suppl. for the PSNR result vs. $t$ on DND.

%
%

\begin{figure}[t]
\begin{center}
\scalebox{1.00}{
\begin{tabular}[t]{c@{ }c@{ }c@{ }c}
\includegraphics[width=.10\textwidth]{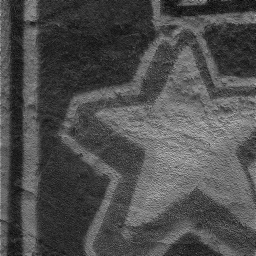}& \hspace{1.2mm}
\includegraphics[width=.10\textwidth]{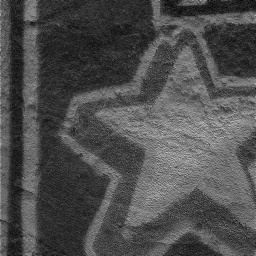}& \hspace{1.2mm}
\includegraphics[width=.10\textwidth]{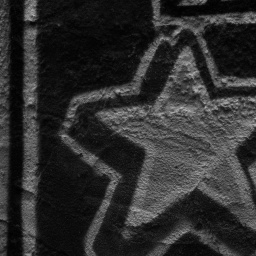}& \hspace{1.2mm}
\includegraphics[width=.10\textwidth]{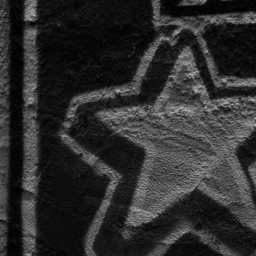}\\
\includegraphics[width=.10\textwidth]{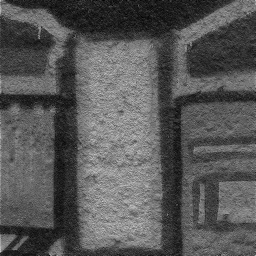}& \hspace{1.2mm}
\includegraphics[width=.10\textwidth]{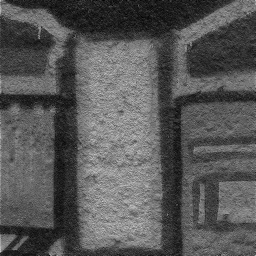}& \hspace{1.2mm}
\includegraphics[width=.10\textwidth]{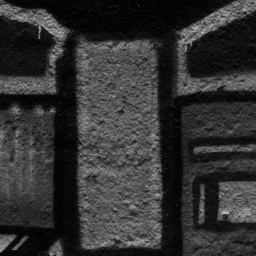}& \hspace{1.2mm}
\includegraphics[width=.10\textwidth]{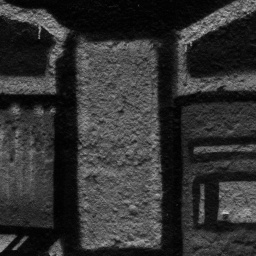}\\
$\mathbf{Y}_{raw}$ &\hspace{1.2mm} $f(\mathbf{Y}^{GT}_{raw})$ & \hspace{1.2mm}$f^{-1}(\mathbf{Y}_{raw})$ & \hspace{1.2mm}$\mathbf{Y}^{GT}_{raw}$

\end{tabular}
}
\end{center}
\vspace*{-2mm}
\caption{Illustration of the assumption $(i)$: there is an invertible element-wise mapping for approximating $\mathbf{Y}^{GT}_{raw}$ with $\mathbf{Y}_{raw}$ and vice versa, \ie, $f(\mathbf{Y}^{GT}_{raw}) \approx \mathbf{Y}_{raw}$ and $f^{-1}(\mathbf{Y}_{raw}) \approx \mathbf{Y}^{GT}_{raw}$.}
\label{fig:raw}
\vspace{-2mm}
\end{figure}
\begin{table}[t]
\begin{center}
\caption{Comparison of Pseudo-ISP and CycleISP for rawRGB denoising on DND~\cite{DND} and SIDD~\cite{SIDD}.}
\label{Table: cycleisp and pesudoisp}
\vspace{-0mm}
\setlength{\tabcolsep}{6pt}
\scalebox{0.60}{
\begin{tabular}{c c c c c}
\toprule
Method&DND&SIDD& \multicolumn{2}{c}{Training Setting} \\ \midrule
CycleISP& 49.12 & 52.41 & \multicolumn{2}{c}{ \textbf{paired} noisy-clean sRGB, paired \textbf{clean} sRGB-rawRGB} \\
Pseudo-ISP& 48.82 & 52.26 & \multicolumn{2}{c}{\textbf{unpaired} noisy-clean sRGB, paired \textbf{noisy} sRGB-rawRGB} \\
\bottomrule
\end{tabular}}
\end{center}\vspace{-8mm}
\end{table}



\vspace*{-1mm}
\subsection{Verifying Assumptions on Noise Modeling}
\label{sec:Assumption_test}
\vspace*{-1mm}
As discussed in Sec.~\ref{sec:noisy_image_generation}, we introduce two assumptions for the learned Pseudo-ISP: $(i)$ invertible element-wise mapping between $\mathbf{Y}^{GT}_{raw}$ and $\mathbf{Y}_{raw}$, and $(ii)$ $\hat{g}(\cdot)$ is a good estimation of $h(\cdot)$.
%
%
%
To verify the assumption $(i)$, we use one patch from DND~\cite{DND} to train the element-wise mapping $f$ and $f^{-1}$ by stacking four $1\!\times\!1$ convolutional layers.
Fig.~\ref{fig:raw} shows $f(\mathbf{Y}^{GT}_{raw})$ and $f^{-1}(\mathbf{Y}_{raw})$ of two other patches from the same image.
Intuitively, both $f(\mathbf{Y}^{GT}_{raw})$ and $f^{-1}(\mathbf{Y}_{raw})$ can respectively well approximate $\mathbf{Y}_{raw}$ and $\mathbf{Y}^{GT}_{raw}$, indicating that the assumption $(i)$ holds on for Pseudo-ISP.


%
%
%

To verify the assumption $(ii)$, we show that it is feasible to learn an effective rawRGB image denoiser by exploiting the element-wise mappings $f$ and $f^{-1}$, and synthetic paired dataset in the pseudo rawRGB space.
Given the learned Pseudo-ISP, we use sRGB2Raw to convert a clean sRGB image to the pseudo rawRGB space, and use Eq.~(\ref{eq:pack_noise}) to synthesize noisy pseudo rawRGB image.
Thus, we constitute a training set to train the same denoising network in~\cite{CycleISP} in the pseudo rawRGB space.
During testing, a noisy rawRGB image is first converted to the pseudo rawRGB space using $f$.
Then, the denoising result is converted to the rawRGB space using $f^{-1}$.
Table~\ref{Table: cycleisp and pesudoisp} lists the results on DND and SIDD rawRGB images. 
It is noteworthy that the training of CycleISP~\cite{CycleISP} requires both paired noisy-clean sRGB and paired sRGB-rawRGB images.
In comparison, we only require unpaired noisy and clean sRGB images for training Pseudo-ISP, and only one pair of noisy sRGB-rawRGB patches for learning $f$ and $f^{-1}$.
The comparable performance of Pseudo-ISP against CycleISP indicates that the learned $\hat{g}(\cdot)$ can serve as a reasonable estimation of $h(\cdot)$.

\vspace*{-1mm}
\subsection{Generalization Ability of Pseudo-ISP}
\label{sec:Generalization_test}
\vspace*{-1mm}
%
%
%


\begin{table}[t]
\centering
\caption{PSNR (dB) results of different color image denoisers on DND~\cite{DND} and SIDD~\cite{SIDD}. Left of $\rightarrow$ is the result of the pre-trained model. Right of $\rightarrow$ corresponds to the result of improved model.}
\label{tab:ab_DND_SIDD}
\setlength{\tabcolsep}{7pt}
\scalebox{0.65}{
\begin{tabular}{c c c}
\toprule
Method&DND&SIDD\\ \midrule
Gaussian Blurring&33.87 $\rightarrow$ 37.53(+3.66)& 28.69 $\rightarrow$ 34.86(+6.17)\\
BM3D~\cite{dabov2008image}&34.51 $\rightarrow$ 37.59(+1.08)& 30.90 $\rightarrow$ 34.91(+4.01)\\
DIP~\cite{DIP}& 36.00 $\rightarrow$ 37.81(+1.81) & 34.21 $\rightarrow$ 35.32(+1.11)\\
CBDNet~\cite{CBDNet}& 38.06 $\rightarrow$ 38.59(+0.53) & 33.26 $\rightarrow$ 34.96(+1.70)\\
RIDNet~\cite{RIDNet}& 39.26 $\rightarrow$ 39.43(+0.17) & 38.70 $\rightarrow$ 38.81(+0.11)\\
PT-MWRN~\cite{cao2020progressive}& 39.84 $\rightarrow$ 40.19(+0.35) & 39.80 $\rightarrow$ 39.92(+0.12)\\
 \bottomrule
\end{tabular}}
\vspace*{-3mm}
\end{table}
\begin{table}[!t]
\begin{center}
\caption{PSNR (dB) results of RIDNet and PT-MWRN on five datasets for assessing Pseudo-ISP in handling noise discrepancy.}
\label{tab:ab_sidd_siddplus}
\setlength{\tabcolsep}{10pt}
\scalebox{0.65}{
\begin{tabular}{c c c c c}
\toprule  Dataset & \multicolumn{2}{c}{RIDNet~\cite{RIDNet}}& \multicolumn{2}{c}{PT-MWRN~\cite{cao2020progressive}}
\\  \midrule
DND&\multicolumn{2}{c}{39.26 $\rightarrow$ 39.43(+0.17)} &\multicolumn{2}{c}{39.84 $\rightarrow$ 40.19(+0.35)}
\\
SIDD&\multicolumn{2}{c}{38.70$\rightarrow$ 38.81(+0.11)} &\multicolumn{2}{c}{39.80 $\rightarrow$ 39.92(+0.12)}
\\ \midrule
SIDDPlus&\multicolumn{2}{c}{36.30 $\rightarrow$ 37.20(+0.90)}&\multicolumn{2}{c}{36.79 $\rightarrow$ 37.35(+0.56)}
\\
\midrule
CC15&\multicolumn{2}{c}{36.83 $\rightarrow$ 37.12(+0.29)} &\multicolumn{2}{c}{36.90 $\rightarrow$ 37.26(+0.36)}
\\
MIT-IP8&\multicolumn{2}{c}{28.16 $\rightarrow$ 28.55(+0.39)}&\multicolumn{2}{c}{28.44 $\rightarrow$ 28.75(+0.31)}
\\
\bottomrule
\end{tabular}}
\vspace*{-10mm}
\end{center}
\end{table}
\begin{figure*}[!t]
\begin{center}
\scalebox{0.84}{
\begin{tabular}{c@{ } c@{ }  c@{ } c@{ }  c@{ } c@{ } c}

    \multirow{4}{*}[+54.7pt]{\includegraphics[width=.31\textwidth,height=.31\textwidth ]{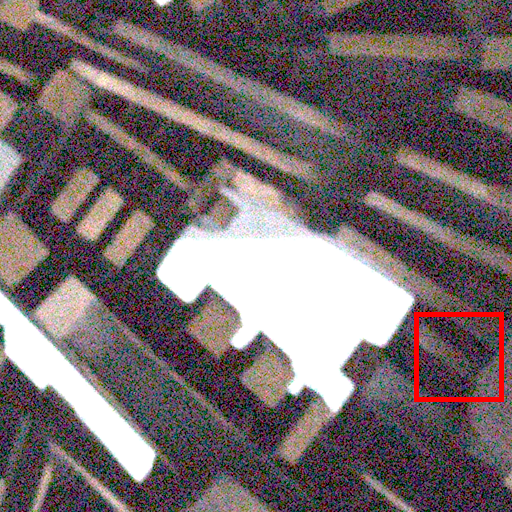}}&
    \includegraphics[width=.128\textwidth]{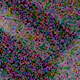}&
    \includegraphics[width=.128\textwidth]{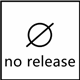}&
  	\includegraphics[width=.128\textwidth]{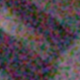}&
    \includegraphics[width=.128\textwidth]{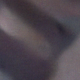}&
 	\includegraphics[width=.128\textwidth]{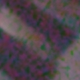}&
  	\includegraphics[width=.128\textwidth]{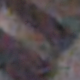}\\
   & \small18.77/0.302 &PSNR/SSIM &\multicolumn{2}{c}{\small24.18/0.500$\rightarrow$ \small32.74/0.887} &\multicolumn{2}{c}{\small27.72/0.680$\rightarrow$ \small30.29/0.790 } \\
   & Noisy&Reference &\multicolumn{2}{c}{Gaussian Blurring} & \multicolumn{2}{c}{DIP~\cite{DIP}} \\

    &
    \includegraphics[width=.128\textwidth]{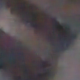}&
    \includegraphics[width=.128\textwidth]{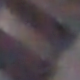}&
    \includegraphics[width=.128\textwidth]{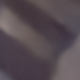}&
    \includegraphics[width=.128\textwidth]{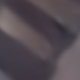}&
    \includegraphics[width=.128\textwidth]{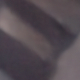}&
    \includegraphics[width=.128\textwidth]{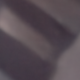}\\
       Original Noisy Image&\multicolumn{2}{c}{\small31.40/0.836 $\rightarrow$ \small32.42/0.866}  &\multicolumn{2}{c}{\small34.30/0.919  $\rightarrow$ \small35.77/0.930} &\multicolumn{2}{c}{\small35.80/0.928 $\rightarrow$ \small36.43/0.936}  \\
     &\multicolumn{2}{c}{CBDNet~\cite{CBDNet}}       &\multicolumn{2}{c}{RIDNet~\cite{RIDNet}}& \multicolumn{2}{c}{PT-MWRN~\cite{cao2020progressive}}\\

\end{tabular}
}
\end{center}
\vspace*{-3mm}
\caption{Denoising results of different methods on real noisy images from DND~\cite{DND}. left:  pre-trained denoiser, right: improved model.}
\label{fig:DND01}
\vspace*{-2.0mm}
\end{figure*}
\begin{table*}[!t]
\begin{center}
\caption{PSNR (dB) and SSIM results of the competing methods on the sRGB images from DND~\cite{DND}, SIDD~\cite{SIDD}, SIDDPlus~\cite{ntire2020}, CC15~\cite{Nam} and MIT-IP8~\cite{MIT}. {\color{red}Red}, {\color{blue}{blue}} and {\color{orange}{orange}} {\color{black}are utilized to indicate top {\color{red}1\textsuperscript{st}}, {\color{blue}2\textsuperscript{nd}} and  {\color{orange}{3\textsuperscript{rd}}} rank, respectively}.}
\label{tab:srgb_dnd_sidd2}
\vspace{-1.5mm}
\setlength{\tabcolsep}{3.3pt}
\scalebox{0.57}{
\begin{tabular}{c c c c c c c c c || c c || c c || c c}
\hline
 Dataset  &GRDN \cite{kim2019grdn}&
 DHDN \cite{park2019densely} & VDN \cite{yue2019variational}&  DANet \cite{DANet}& CycleISP \cite{CycleISP} & DIDN \cite{yu2019deep}& AINDNet \cite{Kim2020Aindnet} & MIRNet~\cite{Zamir2020MIRNet} & CBDNet~\cite{CBDNet}& CBDNet*&RIDNet~\cite{RIDNet} & RIDNet*& PT-MWRN~\cite{cao2020progressive} & PT-MWRN* \\
\hline
DND  & 38.70/0.947  & 39.29/0.952 & 39.38/0.952 & 39.47/0.955 & 39.56/0.956 & 39.62/0.955  &  39.77/\color{red}{0.959}& \color{blue}{39.88}\color{black}{/}0.956 & 38.06/0.898& 38.59/0.946 & 39.26/0.953 & 39.43/0.954 & \color{orange}{39.84}\color{black}{/}\color{orange}{0.958}& \color{red}{40.19}\color{black}{/}\color{red}{0.959}\\

SIDD & \color{blue}{39.85}\color{black}{/}\color{red}{0.959} & \color{orange}{39.84}\color{black}{/}\color{red}{0.959} & 39.27/0.955 & 39.43/0.956 & 39.52/0.957 & 39.78/0.958 &  39.15/0.955 &  39.62/0.958  & 33.26/0.868 & 34.96/0.909 & 38.70/0.950 & 38.81/0.953 & 39.80/\color{red}{0.959} &\color{red}{39.92}\color{black}{/}\color{red}{0.959}\\

\hline
SIDDPlus & 34.51/0.867 & 36.41/0.905 & 36.73/0.917 & 36.86/0.917 & 34.52/0.864 & \color{blue}{36.91}\color{black}{/}0.914 & 36.61/0.909 & \color{orange}{36.87}\color{black}{/}\textcolor{blue}{0.920} & 34.44/0.875 & 35.83/0.886 & 36.30/0.907& 37.20/0.921& 36.79/\color{orange}{0.917} & \color{red}{37.35}\color{black}{/}\color{red}{0.927}\\

\hline
CC15 & 35.39/0.902 & 34.95/0.930 & 35.93/0.941 & \color{blue}{37.20}\color{black}{/}\color{blue}{0.949} & 35.56/0.916 & 36.26/0.945 & 36.12/0.935 & 36.32/0.942 & 36.47/0.939 & 36.99/0.946 & 36.83/0.942& 37.12/0.949& \color{orange}{36.90}\color{black}{/}\color{orange}{0.946} & \color{red}{37.26}\color{black}{/}\color{red}{0.950}\\

MIT-IP8 & 27.05/0.773 & \color{orange}{28.45}\color{black}{/}\color{orange}{0.804} & 28.16/0.779 & 28.20/0.778 & 28.07/0.771 & 28.36/0.790 & 28.22/0.776 & 28.13/0.774 & \color{blue}{28.49}\color{black}{/}\color{blue}{0.812} & 28.64/0.815 & 28.16/0.784& 28.55/0.810 & 28.44/0.802 & \color{red}{28.75}\color{black}{/}\color{red}{0.819}\\

\hline
\end{tabular}}
\vspace{-9.0mm}
\end{center}
\end{table*}
\noindent \textbf{Applying to Different Denoisers.}
We consider three traditional/unsupervised denoisers, \ie, Gaussian blurring, BM3D~\cite{dabov2008image}, DIP~\cite{DIP}, and three deep denoisers, \ie, CBDNet~\cite{CBDNet}, RIDNet~\cite{RIDNet}, PT-MWRN~\cite{cao2020progressive}.
Table~\ref{tab:ab_DND_SIDD} lists the results on DND~\cite{DND} and SIDD~\cite{SIDD}, and we have the following observations:
$(i)$ Pseudo-ISP can be applied to different denoisers for boosting performance.
$(ii)$ More significant improvements can be got for traditional/unsupervised denoisers not specified for real-world noisy photographs.
$(iii)$ Albeit RIDNet and PT-MWRN are pre-trained with SIDD training, their performance can also be improved on SIDD testing.

\noindent \textbf{Handling Different Kinds of Noise Discrepancy.}
Using RIDNet~\cite{RIDNet} and PT-MWRN~\cite{cao2020progressive} pre-trained on SIDD training, we assess the ability of Pseudo-ISP in handling three kinds of noise discrepancy.
We consider five datasets.
%
DND and SIDD have the similar noise characteristics with SIDD training, and thus the noise discrepancy is small.
As an extension for NTIRE2020 challenge, the noise characteristics of SIDDPlus differs from SIDD training, resulting in large noise discrepancy.
Albeit the noise discrepancy is large for CC15 and MIT-IP8, the images from these two datasets are JPEG compressed, increasing the difficulty of Pseudo-ISP learning.
Table~\ref{tab:ab_sidd_siddplus} lists the results on the five datasets.
For DND and SIDD, the gains by Pseudo-ISP are moderate (\ie, 0.1$ \sim $0.2 dB) due to small noise discrepancy.
For SIDDPlus, the PSNR gains are notable (\ie, $>$ 0.5 dB), owing to the ability of Pseudo-ISP in alleviating noise discrepancy.
For CC15 and MIT-IP8, JPEG compression and complex demosaicking algorithm (MIT-IP8) limit the effectiveness of Pseudo-ISP.
Nonetheless, Pseudo-ISP can still achieve PSNR gains of 0.3$ \sim $0.4 dB, indicating its generalization ability in handling noise discrepancy.

%

%

%

{\color{red}
}

\noindent \textbf{Comparison with Other Adaption Methods.}
%
%
We compare Pseudo-ISP with two baselines by $(i)$ finetuning pre-trained denoiser with its original training data for extra 50 epochs, and $(ii)$ incorporating rotation/flip augmentation and finetuning denoiser using pseudo paired images.
%
%
The results of CBDNet and RIDNet on DND are given in the suppl.
The baselines bring very limited improvement (\ie, $<$ 0.05 dB) in comparison to Pseudo-ISP (0.53/0.17 dB for CBDNet/RIDNet).
So the gain of Pseudo-ISP should be ascribed to denoiser adaption instead of increasing training time.

\vspace*{-2mm}
\subsection{Comparison with State-of-the-arts}
\label{sec:Comparison_SOTA}
\vspace*{-1mm}
We apply Pseudo-ISP for adapting CBDNet, RIDNet and PT-MWRN (\ie, CBDNet*, RIDNet* and PT-MWRN*), and compare them with 11 state-of-the-art denoisers on five datasets. 
Table~\ref{tab:srgb_dnd_sidd2} lists the PSNR and SSIM results.
On all datasets, CBDNet*, RIDNet* and PT-MWRN* outperform their counterparts, indicating that our Pseudo-ISP can be incorporated with different pre-trained denoisers for handling various kinds of noise discrepancy.
Moreover, PT-MWRN* achieves the best quantitative performance on the five datasets.
On SIDDPlus, PT-MWRN* outperforms the second best competing method, \ie, DIDN~\cite{yu2019deep}, by a large margin of 0.44 dB, owing to the large noise discrepancy between SIDDPlus and original training set.

%
%
%
%
%
%

Fig.~\ref{fig:DND01} shows the visualized comparison by incorporating Pseudo-ISP with different denoisers on DND.
More results on other datasets can be found in the suppl.
For traditional and unsupervised methods, Pseudo-ISP can improve the visual quality obviously.
On CC15, the improvement by Pseudo-ISP is visually perceivable even for deep denoisers, \eg, CBDNet, RIDNet and PT-MWRN (see the suppl.).
Once denoiser adaption is done, Pseudo-ISP improves denoising performance without bringing additional computation cost (see the suppl.), further making it very competitive.

%
%
%
%
%
%


%
%

\vspace*{-3mm}
\section{Conclusion}
\vspace*{-2mm}
In this work, we presented an unpaired learning scheme which alternates between Pseudo-ISP learning and denoiser adaption by using a pre-trained denoiser, a set of test noisy images and an unpaired set of clean images.
Pseudo-ISP is introduced for noise modeling to synthesize realistic noisy images.
By re-training the pre-trained model using both pseudo and synthetic pairs, existing denoisers can then be adapted to handle noisy discrepancy.
Experimental results show that our method is effective in boosting existing denoisers to adapt to real-world noisy image datasets.
In the future, we will extend Pseudo-ISP for more challenging and precise image noise modeling, \eg, low-light image noise.

{\small
\bibliographystyle{ieee_fullname}
\bibliography{egbib}
}

\renewcommand\thesection{\Alph{section}}
\renewcommand\thefigure{\Alph{figure}}
\renewcommand\thetable{\Alph{table}}
\renewcommand\thesubsection{\thesection.\arabic{subsection}}

\newcommand*{\affaddr}[1]{#1} 
\newcommand*{\affmark}[1][*]{\textsuperscript{#1}}
\newcommand*{\email}[1]{\texttt{#1}}

\newcommand{\tabincell}[2]{\begin{tabular}{@{}#1@{}}#2\end{tabular}}
\twocolumn[
\begin{center}
	{\LARGE \textbf{Supplemental Materials\\~\\~\\}}
\end{center}]
\setcounter{section}{0}
\setcounter{table}{0}
\setcounter{figure}{0}

\hspace{-5mm}The content of this supplementary material involves:
\vspace{2mm}

	\hspace{-4mm}A. Illustration of Noise Discrepancy in Sec.~\ref{sec:Illustratio_of_Noise Discrepancy}.

	\hspace{-4mm}B. Derivation of Eq~(\ref{eq:loss_noise}) in Sec.~\ref{sec:Derivation_of_Loss_Function}.

	\hspace{-4mm}C. Difference between Pseudo-ISP and CycleISP in Sec.~\ref{sec:Difference_between}.

    \hspace{-4mm}D. More Ablation Studies in Sec.~\ref{sec:More_Ablation_Studies}.

	\hspace{-4mm}E. More Quantization and Qualitative Results in Sec.~\ref{sec:More_Quantization_and_Qualitative_Results}.

\vspace{2mm}

\section{Illustration of Noise Discrepancy}
\label{sec:Illustratio_of_Noise Discrepancy}
In this section, we first test the denoising performance of the same model on two validation sets with different noise distributions, which is evaluated each epoch during the whole training period.
Then, we apply the proposed unpaired learning scheme to pre-trained denoiser on these two validation sets.

\begin{figure}[t]
\scriptsize{
\begin{center}
\begin{overpic}[width=0.40\textwidth,scale=0.5]{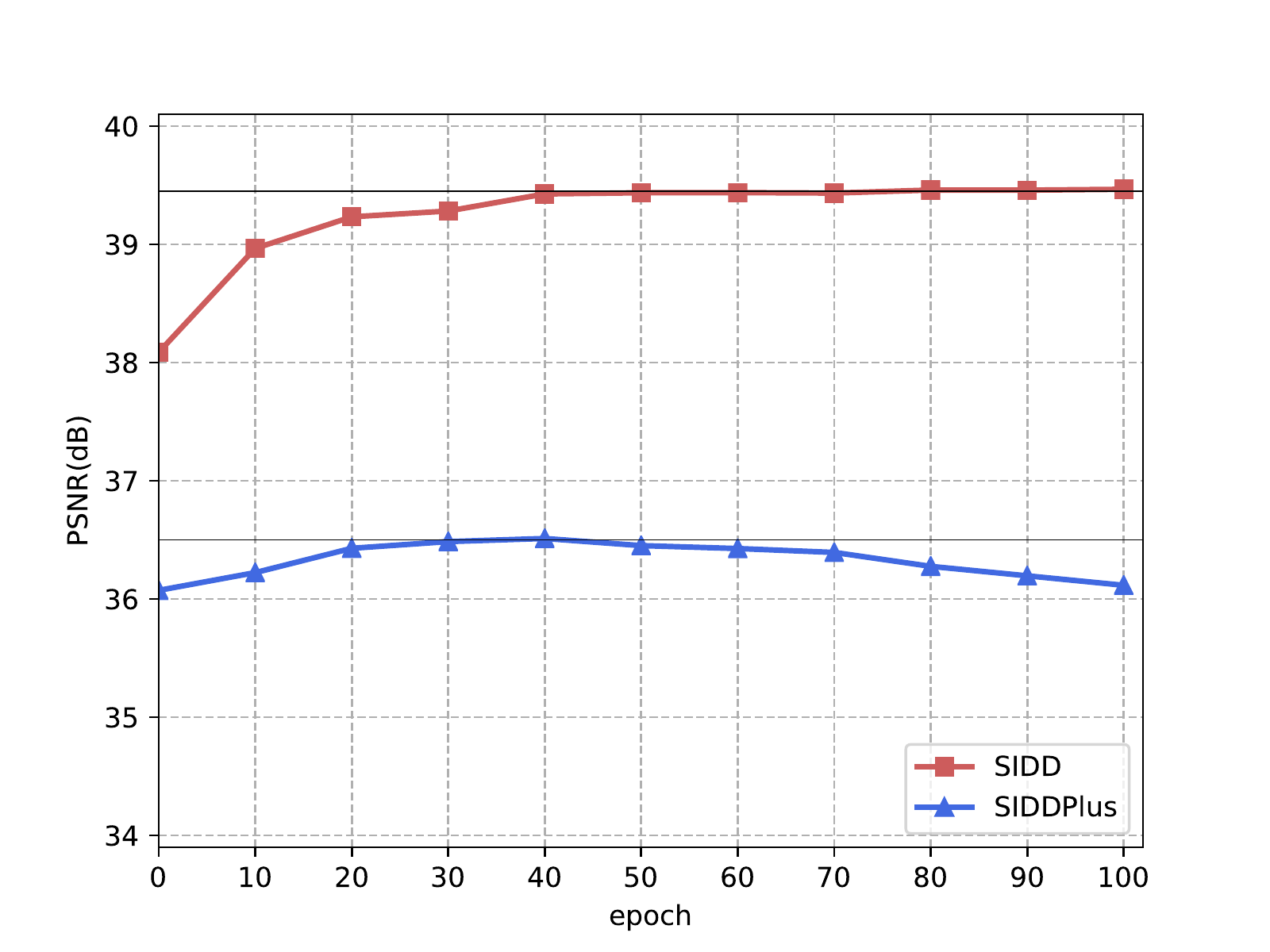} 
\put(15,62.8){\color{black}{\tiny 39.47}}
\put(15,34.7){\color{black}{\tiny 36.57}}
\end{overpic}
\caption{\small
{PSNR  (dB) curves of MWCNN~\cite{MWCNN} on SIDD validation and SIDDPlus validation dataset. }}
\label{fig:chart}
\vspace{-2mm}
\end{center}}
\end{figure}

To illustrate the noise discrepancy clearly, we perform extensive experiments on SIDD validation and SIDDPlus validation set.
We train MWCNN~\cite{MWCNN} using SIDD~\cite{SIDD} training set for 100 epochs and evaluate each epoch on SIDD validation and SIDDPlus validation set.
As an extension for NTIRE2020 challenge, the noisy images in validation set of SIDDPlus differs from images in training set of SIDD.
PSNR curve results are presented in Fig.~\ref{fig:chart}.
On the one hand, MWCNN~\cite{MWCNN} presents better performance on SIDD validation set than SIDDPlus validation set.
This is mainly because the SIDD training set and the SIDD validation set are in a relatively close noise distribution, but the SIDDPlus validation set is inconsistent with their distribution.
On the other hand, PSNR of SIDD validation set increases gradually with the continuous training process and tends to be stable after about 40 epochs.
However, result of SIDDPlus validation set decreases after 40 epochs.
The main reason for performance drop is that the denoiser is over-fitted to the specific noise distribution on SIDD training set, and exhibits poor generalization ability on SIDDPlus validation set with a different noise distribution, \ie, noise discrepancy.
To tackle the noise discrepancy issue, we present an unpaired learning scheme to adapt a color image denoiser for handling test images with noise discrepancy.
We evaluate on SIDD validation set and SIDDPlus validation set with two denoisers, \ie, MWCNN~\cite{MWCNN} and RIDNet~\cite{RIDNet}.
Both models are pre-trained using SIDD training set.
From Table~\ref{tab:s_sidd_siddplus}, the pre-trained MWCNN~\cite{MWCNN} and RIDNet~\cite{RIDNet} overfit to the SIDD training data, in which the noisy images are consistent with SIDD validation set, but show poor generalization ability on SIDDPlus validation images.
Although there is only a little improvement (about $0.1$dB) on the SIDD validation set, it also shows that the proposed Pseudo-ISP noise model can generate consistent distributed noisy images.
Nonetheless, this noise discrepancy issue can be largely mitigated by our unpaired learning scheme.
Benefited from denoiser adaption, the performance on SIDDPlus validation set can be significantly improved (\ie, $>0.9$dB) in comparison to the pre-trained counterparts.

\begin{table}[t]
\begin{center}
\caption{PSNR (dB) results of of MWCNN~\cite{MWCNN} and RIDNet~\cite{RIDNet} on SIDD Validation set (SIDD Val.) and SIDDPlus Validation set (SIDDPlus Val.).}
\label{tab:s_sidd_siddplus}
\setlength{\tabcolsep}{11pt}
\scalebox{0.80}{
\begin{tabular}{c c c c c}
\toprule  Dataset & \multicolumn{2}{c}{MWCNN~\cite{MWCNN}}& \multicolumn{2}{c}{RIDNet~\cite{RIDNet}}
\\  \midrule
SIDD Val. &\multicolumn{2}{c}{39.47$\rightarrow$39.58(+0.11)}&\multicolumn{2}{c}{38.71$\rightarrow$38.80(+0.09)}
\\
SIDDPlus Val. &\multicolumn{2}{c}{36.10$\rightarrow$37.05(+0.95)}&\multicolumn{2}{c}{36.30$\rightarrow$37.20(+0.90)}
\\
\bottomrule
\end{tabular}}
\vspace{-8mm}
\end{center}
\end{table}

\section{Derivation of the Eq~(\ref{eq:loss_noise})}
\label{sec:Derivation_of_Loss_Function}

We elaborate on the loss function about Eq~(\ref{eq:loss_noise}).
Since the ground-truth rawRGB space noise is assumed to be signal-dependent and spatially independent, the ground-truth noisy rawRGB image $\mathbf{Y}^{GT}_{raw}$ and the corresponding ground-truth clean one $\mathbf{X}^{GT}_{raw}$ at pixel $i$ then can be written as:

\begin{equation}
\setlength{\abovedisplayskip}{5pt}
\setlength{\belowdisplayskip}{5pt}
\label{eq:gt_noisy_n0_i}
\mathbf{Y}^{GT}_{raw}[i] = \mathbf{X}^{GT}_{raw}[i] + \bm{\sigma}[i] \cdot \mathbf{n}_{0}[i],
\end{equation}
where $\bm{\sigma}$ denotes standard deviation of ground-truth rawRGB space noise, and $\mathbf{n}_{0}$ is the sampling noise following the standard normal distribution.
We exploit the entry-wise absolute term ${| \mathbf{Y}^{GT}_{raw} \!-\! \mathbf{X}^{GT}_{raw} |}$ to help noise estimation subnet learn the noise level $\bm{\sigma}$.
The term ${| \mathbf{Y}^{GT}_{raw} \!-\! \mathbf{X}^{GT}_{raw} |}$ obeys folded normal distribution \cite{leone1961folded}.
So the mean of this term:
\begin{equation}
\label{eq:mean}
\small
\begin{split}
\mu_{| \mathbf{Y}^{GT}_{raw} \!-\! \mathbf{X}^{GT}_{raw}|}=\bm{\sigma} \sqrt{\frac{2}{\pi}} e^{-\frac{{\bm{\mu}}^{2}}{2 {\bm{\sigma}}^{2}}}-{\bm{\mu}}\left(1-2 \Phi\left(\frac{{\bm{\mu}}}{\bm{\sigma}}\right)\right)
\end{split}
\end{equation}
where $\Phi$ is the normal cumulative distribution function, and $\bm{\mu}$ denotes the mean of ground-truth rawRGB space noise.
Under the general signal-dependent and spatially independent ground-truth rawRGB space assumption with $\bm{\mu} = 0$, $\mu_{| \mathbf{Y}^{GT}_{raw} \!-\! \mathbf{X}^{GT}_{raw}|}=\sqrt{\frac{2}{\pi}} \bm{\sigma}$.
Therefore, for the noise estimation, we utilize $\sqrt{\frac{\pi}{2}}| \mathbf{Y}^{GT}_{raw} \!-\! \mathbf{X}^{GT}_{raw}|$ as supervision for joint training.

\section{Difference between Pseudo-ISP and CycleISP}
\label{sec:Difference_between}

Difference between Pseudo-ISP and CycleISP~\cite{CycleISP} can be summarized from two aspects:
$(i)$ Despite learning agnostic-ISP pipeline, our approach differs from CycleISP~\cite{CycleISP} definitely.
CycleISP~\cite{CycleISP} aims to produce realistic image pairs by learning ISP pipeline in a data-driven manner, which overly depends on numerous paired clean-noisy and sRGB-rawRGB images. It performance degrades dramatically when applied to images with noise discrepancy.
Our Pseudo-ISP is mainly designed to synthesize noisy images adaptive to the domain of test noisy image in sRGB space. The synthetic paired data are then used to re-train the denoiser to address the noise discrepancy issue.
$(ii)$ CycleISP~\cite{CycleISP} ignores noise model, while the characteristics of rawRGB images noise are effectively captured by the proposed noise estimation subnet in Pseudo-ISP.
Furthermore, our Pseudo-ISP can be trained in an end-to-end way, while the complex architecture of CycleISP~\cite{CycleISP} need to be trained by multiple steps.

To further verify the performance of Pseudo-ISP, we provide the model parameters and running time (noisy image generation time) comparison in Table~\ref{tab:s_ab_loss_function}.
Notice that the parameters of CycleISP~\cite{CycleISP} are 6 times that of Pseudo-ISP.
As for the noisy image generation, the Pseudo-ISP is 3 times faster than CycleISP~\cite{CycleISP}.
Obviously, Pseudo-ISP can achieve a good balance between model performance, parameters, and running time, which provide a lightweight noise model.

\begin{table}[!t]
\begin{center}
\caption{Study on the model parameters and running time (the dimension of the test image is $256 \times 256$) between CycleISP and Pseudo-ISP.}
\label{tab:cy and pe}
\vspace{+2mm}
\setlength{\tabcolsep}{14pt}
\scalebox{0.90}{
\begin{tabular}{c c c}
\toprule
Model  &CycleISP& Pseudo-ISP\\
\hline
Parameters($10^6$) & 7.47 & 1.25 \\
Time(ms) & 83.9 & 27.9 \\
\bottomrule
\end{tabular}}
\end{center}\vspace{-1.4em}
\end{table}

\section{More Ablation Studies}
\label{sec:More_Ablation_Studies}

In this section, we conduct detailed ablation studies of Pseudo-ISP, including different loss functions, times of alternated training and comparison with other adaption methods.

\begin{table}[t]
\begin{center}
\caption{Ablation study for different supervisions for noise estimation subnet. PSNR (dB) results on DND~\cite{DND}.}
\label{tab:s_ab_loss_function}
\setlength{\tabcolsep}{11pt}
\scalebox{0.80}{
\begin{tabular}{c c c}
\toprule loss function & Eq~(\ref{eq:loss_pseudoisp}) & Eq~(\ref{eq:loss_pseudoisp_2}) \\ \midrule
PSNR  & 38.06 $\rightarrow$ 38.59 (+0.53) & 38.06 $\rightarrow$ 38.40 (+0.34) \\ \bottomrule
\end{tabular}}
\vspace{-4mm}
\end{center}
\end{table}

\noindent \textbf{Different Loss Functions.}
As detailed derivation in~\ref{sec:Derivation_of_Loss_Function}, the mean of term $(\mathbf{Y}^{GT}_{raw} \!-\! \mathbf{X}^{GT}_{raw} )^{2}$ is $\bm{\sigma^2}$.
So we then utilize this term as supervision for learning noise model.
Thus, the loss function for training pseudo ISP and pseudo noise model is changed as follow:

\begin{equation}
\setlength{\abovedisplayskip}{5pt}
\setlength{\belowdisplayskip}{5pt}
\label{eq:loss_pseudoisp_2}
\small
\begin{split}
\!\!\!\!\!\!\!
\mathcal{L}_{P} \!\!=\!\!
\Big \| \hat{\mathbf{X}}^{*} \!\!-\!\! \hat{\mathbf{X}}  \Big \|^{2}
\!+\! \Big \| \mathbf{Y}^{*} \!\!-\!\! \mathbf{Y}  \Big \|^{2}
\!+\!\lambda \Big \| \bm{\hat{\sigma}^{2}} \!-\! (\mathbf{Y}_{\text {pack}} \!-\! \mathbf{\hat{X}}_{\text {pack}} )^2 \Big \| ^{2}
\end{split}
\end{equation}
We conduct experiments with different supervision for noise estimation subnet.
We select CBDNet~\cite{CBDNet} as the baseline denoiser for evaluation on DND~\cite{DND} dataset.
From Table~\ref{tab:s_ab_loss_function}, using $\sqrt{\frac{\pi}{2}} | \mathbf{Y}_{\text {pack}} \!-\! \mathbf{\hat{X}}_{\text {pack}}|$ as supervision for noise estimation subnet can effectively learn more accurate noise level.
\begin{figure}[t]
\scriptsize{
\begin{center}
\vspace{-0ex}
\begin{overpic}[width=0.40\textwidth,scale=0.7]{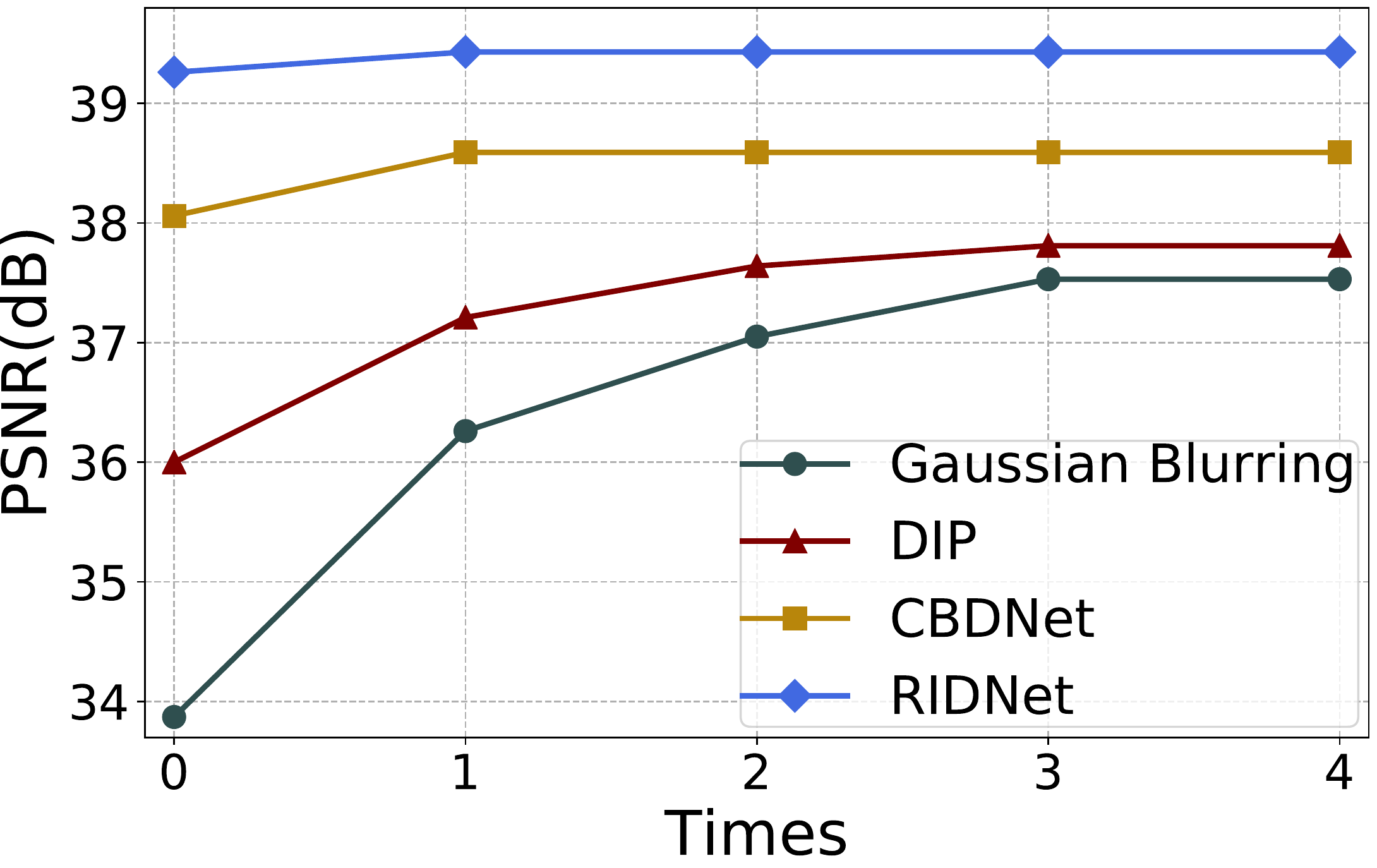} 
\end{overpic}
\end{center}}
\vspace{-4.0ex}
\caption{\small
{PNSR (dB) results for varying times of alternated training (\ie, $t$).}
}
\label{fig:DND_SIDD_chart}
\vspace*{+2mm}
\end{figure}

\noindent \textbf{Times of Alternated Training.}
Fig.~\ref{fig:DND_SIDD_chart} shows the PNSR values obtained using different times of alternated training (\ie, $t$).
It can be seen that increasing the times of alternated training continuously improves the denoising performance, and the gains becomes negligible when $t \geq 3$.
%
\begin{table}[!t]\footnotesize
\centering
\vspace{-0.12in}
\caption{
Comparison of our Pseudo-ISP with other finetuning/adaption methods on DND~\cite{DND}.
}
\scalebox{0.85}{
\renewcommand\tabcolsep{1.0pt}
\begin{tabular}{p{3.5cm}<{\centering} p{2.8cm}<{\centering} p{2.8cm}<{\centering}}
\toprule Adaption Method & CBDNet~\cite{CBDNet} & RIDNet~\cite{RIDNet} \\ \midrule
Finetune-I &  38.06$\rightarrow$38.09 (+0.03)&  39.26$\rightarrow$39.30 (+0.04)\\
Finetune-II&  38.06$\rightarrow$38.10 (+0.04)&  39.26$\rightarrow$39.28 (+0.02)\\
Pseudo-ISP&  38.06$\rightarrow$\textbf{38.59} (+\textbf{0.53})& 39.26$\rightarrow$ \textbf{39.43} (+\textbf{0.17})\\ \bottomrule
\end{tabular}
}
\label{tab:adapt_methods}
\vspace{-0.15in}
\end{table}

\noindent \textbf{Comparison with Other Adaption Methods.}
Pseudo-ISP leverages additional training time for denoiser adaption.
Thus, we compare Pseudo-ISP with two baselines by increasing training time to pre-trained denoiser in Table~\ref{tab:adapt_methods}.
For Finetune-I, we finetune pre-trained denoiser with its original training data for extra 50 epochs.
For Finetune-II, we incorporate rotation or/and flip based data augmentation and finetune pre-trained denoiser using the pseudo paired images.
Table~\ref{tab:adapt_methods} lists the results of CBDNet and RIDNet on DND.
Finetune-I and Finetune-II bring very limited improvement (\ie, $< 0.05$ dB).
While the PSNR gains by Pseudo-ISP are 0.53 dB and 0.17 dB for CBDNet and RIDNet, respectively.
Thus, the effectiveness of Pseudo-ISP can be ascribed to denoiser adaption instead of the increase of training time.

\section{More Quantization and Qualitative Results}
\label{sec:More_Quantization_and_Qualitative_Results}
\begin{figure*}
\begin{center}
\scalebox{0.84}{
\begin{tabular}{c@{ } c@{ }  c@{ } c@{ }  c@{ } c@{ } c}

    \multirow{4}{*}[+54.7pt]{\includegraphics[width=.31\textwidth,height=.31\textwidth ]{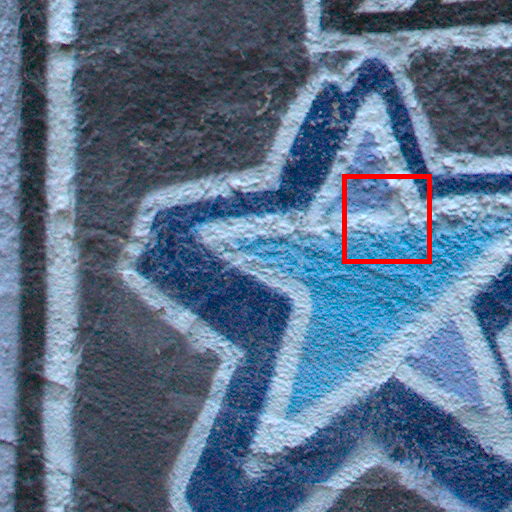}}&
    \includegraphics[width=.128\textwidth]{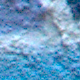}&
    \includegraphics[width=.128\textwidth]{fig/no}&
  	\includegraphics[width=.128\textwidth]{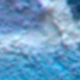}&
    \includegraphics[width=.128\textwidth]{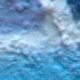}&
 	\includegraphics[width=.128\textwidth]{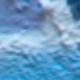}&
  	\includegraphics[width=.128\textwidth]{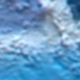}\\
    &\small28.48/0.901  &PSNR/SSIM    & \multicolumn{2}{c}{\small30.37/0.940$\rightarrow$ \small31.73/0.959} & \multicolumn{2}{c}{\small 29.87/0.931$\rightarrow$  \small30.86/0.957 } \\
    & Noisy&Reference &\multicolumn{2}{c}{Gaussian Blurring} & \multicolumn{2}{c}{DIP~\cite{DIP}}\\

    &
    \includegraphics[width=.128\textwidth]{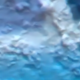}&
    \includegraphics[width=.128\textwidth]{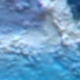}&
    \includegraphics[width=.128\textwidth]{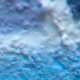}&
    \includegraphics[width=.128\textwidth]{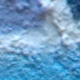}&
    \includegraphics[width=.128\textwidth]{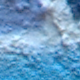}&
    \includegraphics[width=.128\textwidth]{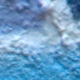}\\
      Original Noisy Image &\multicolumn{2}{c}{\small31.06/0.955 $\rightarrow$ \small31.37/0.959}  &\multicolumn{2}{c}{\small32.31/0.964  $\rightarrow$ \small32.73/0.969} & \multicolumn{2}{c}{\small32.68/0.968 $\rightarrow$ \small33.10/0.971} \\
    &\multicolumn{2}{c}{CBDNet~\cite{CBDNet}}       &\multicolumn{2}{c}{RIDNet~\cite{RIDNet}}& \multicolumn{2}{c}{PT-MWRN~\cite{cao2020progressive}}\\

\end{tabular}
}
\end{center}
\vspace*{-4mm}
\caption{Denoising results of different methods on real noisy images from DND~\cite{DND}.}
\label{fig:DND03}
\vspace*{-4mm}
\end{figure*}

\begin{figure*}[t]
\begin{center}
\scalebox{0.97}{
\begin{tabular}{c@{ } c@{ }  c@{ } c@{ }  c@{ } c}\hspace{-2mm}
    \includegraphics[width=.155\textwidth]{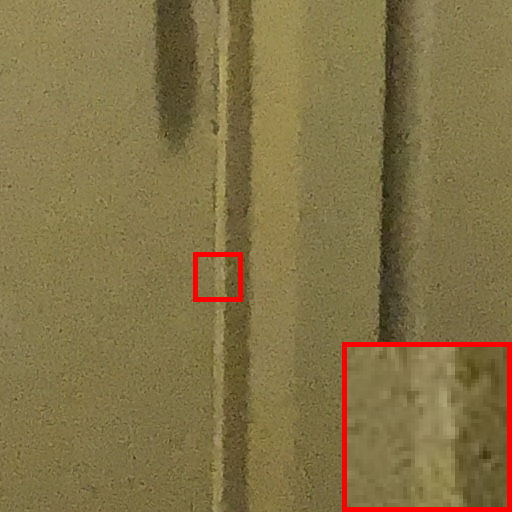}&\hspace{-1.2mm}
    \includegraphics[width=.155\textwidth]{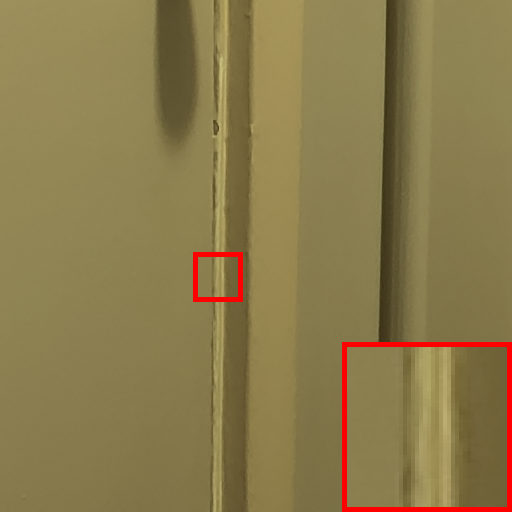}&\hspace{-1.2mm}
  	\includegraphics[width=.155\textwidth]{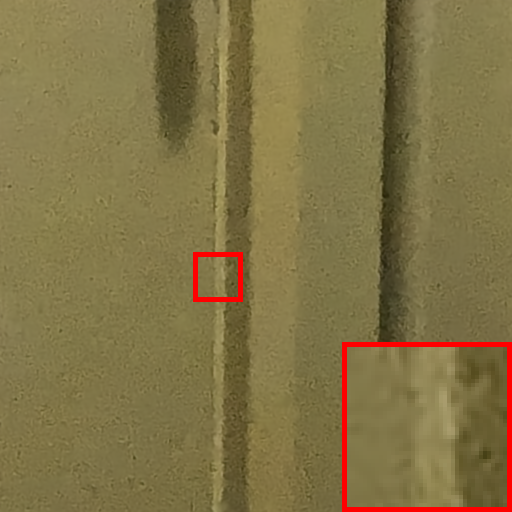}&\hspace{-1.2mm}
    \includegraphics[width=.155\textwidth]{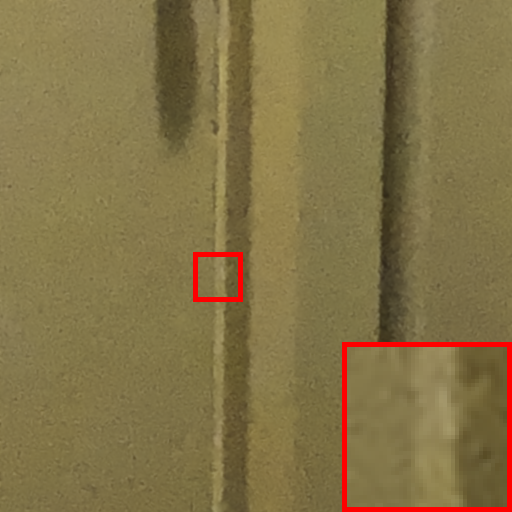}&\hspace{-1.2mm}
 	\includegraphics[width=.155\textwidth]{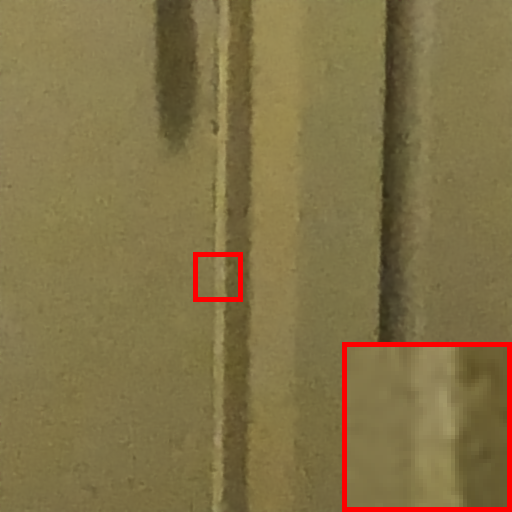}&\hspace{-1.2mm}
  	\includegraphics[width=.155\textwidth]{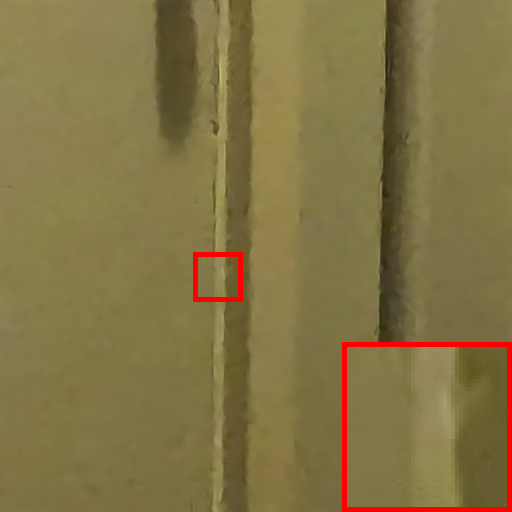}\\
  27.73/0.656 & PSNR/SSIM & \multicolumn{2}{c}{28.51/0.813 $\rightarrow$ 29.30/0.860} & \multicolumn{2}{c}{29.32/0.919 $\rightarrow$ 29.50/0.945} \\
  Noisy & Reference & \multicolumn{2}{c}{RIDNet~\cite{RIDNet}} & \multicolumn{2}{c}{PT-MWRN~\cite{cao2020progressive}}
\end{tabular}
}
\end{center}\vspace*{-3mm}
\caption{Denoising results of different methods on real noisy images from MIT-IP8~\cite{MIT}.}
\label{Fig:MIT-IP8}
\end{figure*}

\begin{figure*}[t]
\scriptsize{
\begin{center}
\subfigure{
\begin{minipage}[c]{0.22\textwidth}
\centering
  \includegraphics[width=1\linewidth]{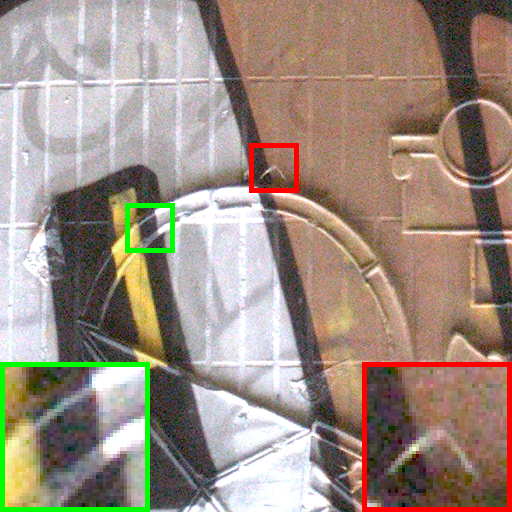}
  \centerline{\scriptsize 26.90/0.753}
  \centerline{\scriptsize Noisy}
\end{minipage}%
}
\subfigure{
\begin{minipage}[c]{0.22\textwidth}
\centering
  \includegraphics[width=1\linewidth]{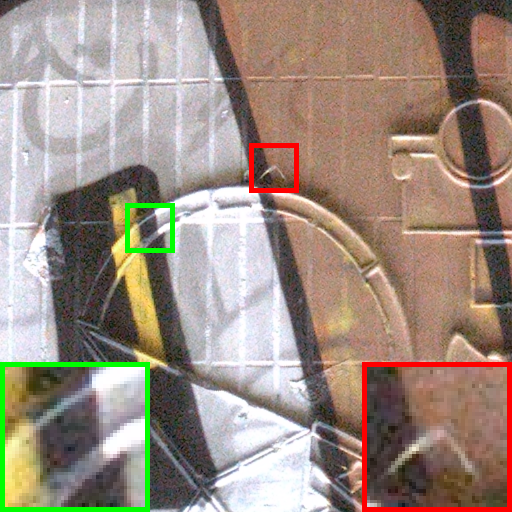}
  \centerline{\scriptsize 28.95/0.818}
  \centerline{\scriptsize CDnCNN-B\cite{DnCNN}}
\end{minipage}%
}
\subfigure{
\begin{minipage}[c]{0.22\textwidth}
\centering
  \includegraphics[width=1\linewidth]{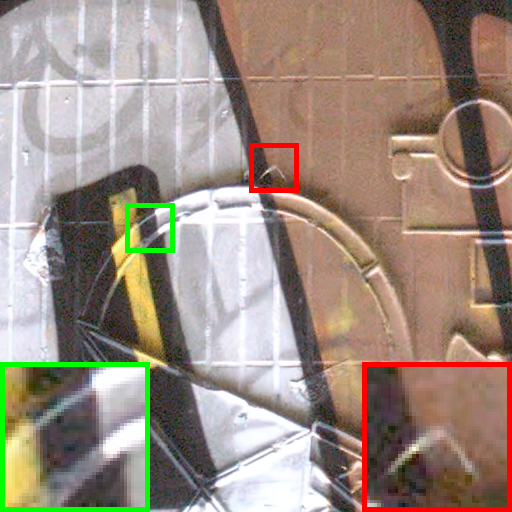}
  \centerline{\scriptsize 30.91/0.872}
  \centerline{\scriptsize BM3D \cite{dabov2008image}}
\end{minipage}%
}
\subfigure{
\begin{minipage}[c]{0.22\textwidth}
\centering
  \includegraphics[width=1\linewidth]{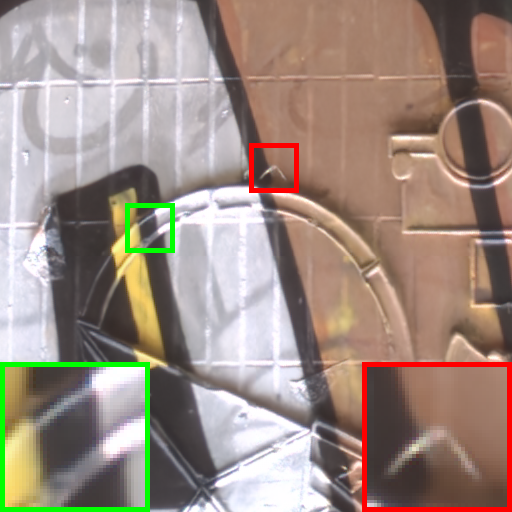}
  \centerline{\scriptsize 34.30/0.940}
  \centerline{\scriptsize GRDN \cite{kim2019grdn}}
\end{minipage}%
}
\\
\vspace{-3ex}
\subfigure{
\begin{minipage}[c]{0.22\textwidth}
\centering
  \includegraphics[width=1\linewidth]{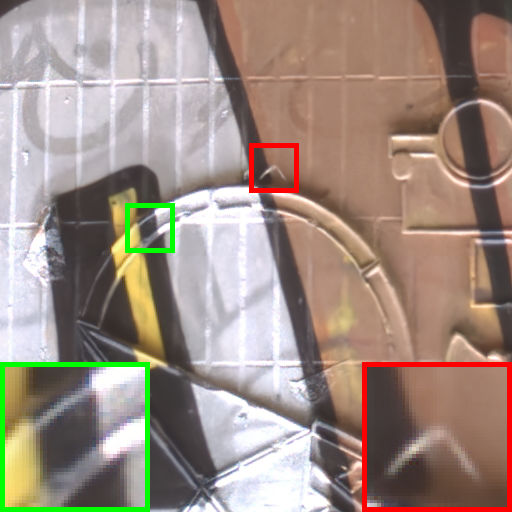}
  \centerline{\scriptsize 34.57/0.942}
  \centerline{\scriptsize DHDN \cite{park2019densely}}
\end{minipage}%
}
\subfigure{
\begin{minipage}[c]{0.22\textwidth}
\centering
  \includegraphics[width=1\linewidth]{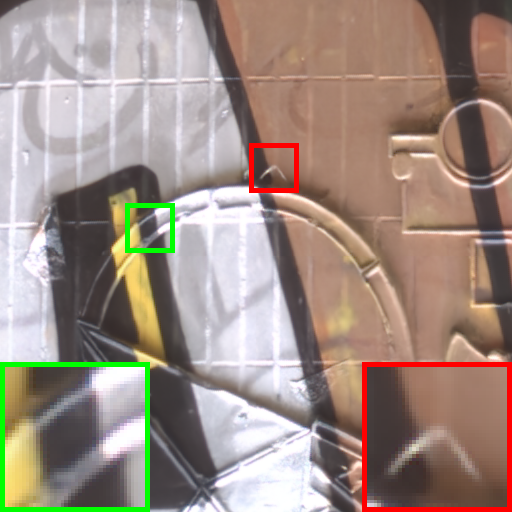}
  \centerline{\scriptsize 34.24/0.941}
  \centerline{\scriptsize DIDN \cite{yu2019deep}}
\end{minipage}%
}
\subfigure{
\begin{minipage}[c]{0.22\textwidth}
\centering
  \includegraphics[width=1\linewidth]{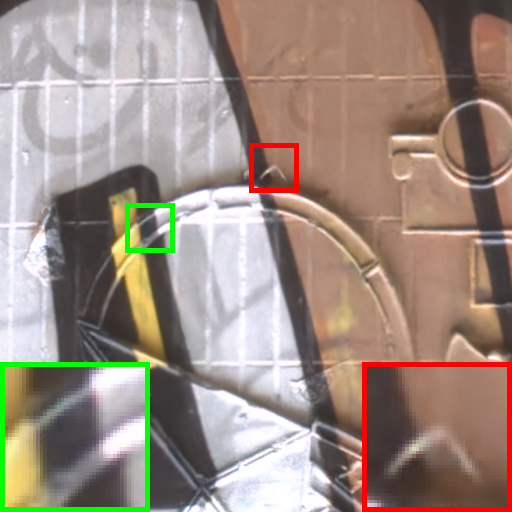}
  \centerline{\scriptsize 33.89/0.938}
  \centerline{\scriptsize VDN \cite{yue2019variational}}
\end{minipage}%
}
\subfigure{
\begin{minipage}[c]{0.22\textwidth}
\centering
  \includegraphics[width=1\linewidth]{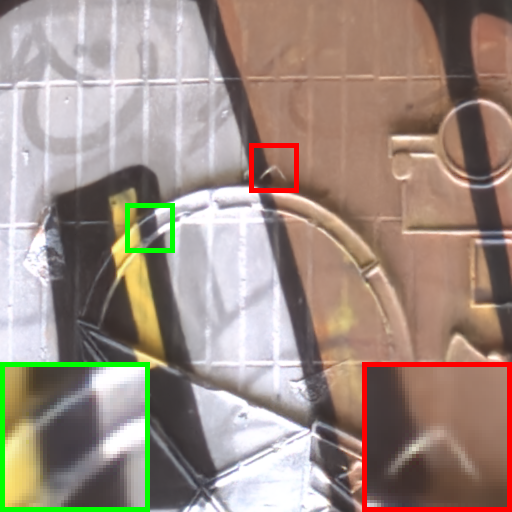}
  \centerline{\scriptsize 34.36/0.941}
  \centerline{\scriptsize DANet \cite{DANet}}
\end{minipage}%
}
\\
\vspace{-3ex}
\subfigure{
\begin{minipage}[c]{0.22\textwidth}
\centering
  \includegraphics[width=1\linewidth]{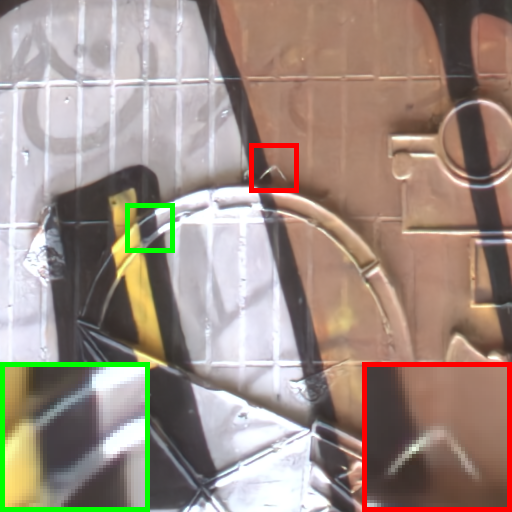}
  \centerline{\scriptsize 34.32/0.941}
  \centerline{\scriptsize CycleISP \cite{CycleISP}}
\end{minipage}%
}
\subfigure{
\begin{minipage}[c]{0.22\textwidth}
\centering
  \includegraphics[width=1\linewidth]{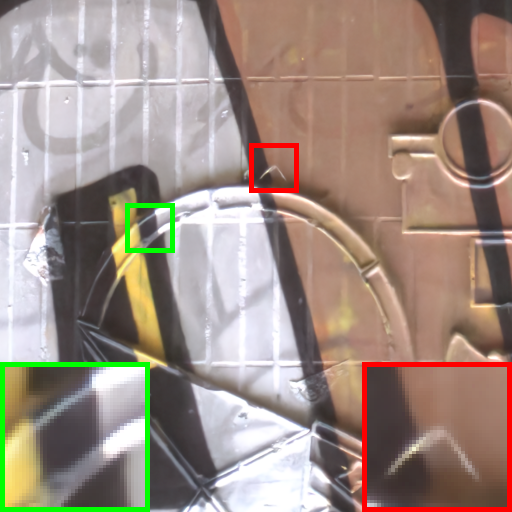}
  \centerline{\scriptsize 34.36/0.943}
  \centerline{\scriptsize AINDNet \cite{Kim2020Aindnet}}

\end{minipage}%
}
\subfigure{
\begin{minipage}[c]{0.22\textwidth}
\centering
  \includegraphics[width=1\linewidth]{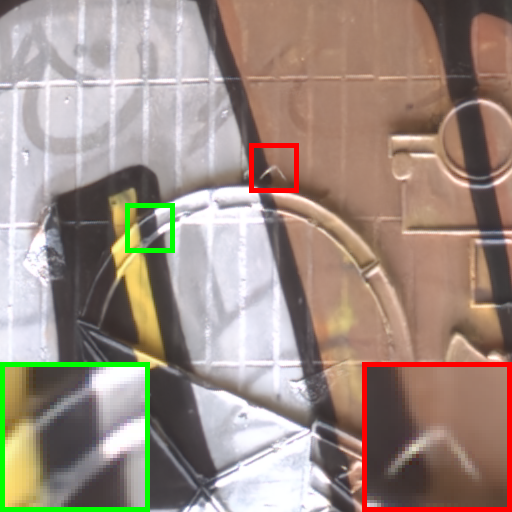}
  \centerline{\scriptsize 34.77/0.944}
  \centerline{\scriptsize MIRNet ~\cite{Zamir2020MIRNet}}

\end{minipage}%
}
\subfigure{
\begin{minipage}[c]{0.22\textwidth}
\centering
  \includegraphics[width=1\linewidth]{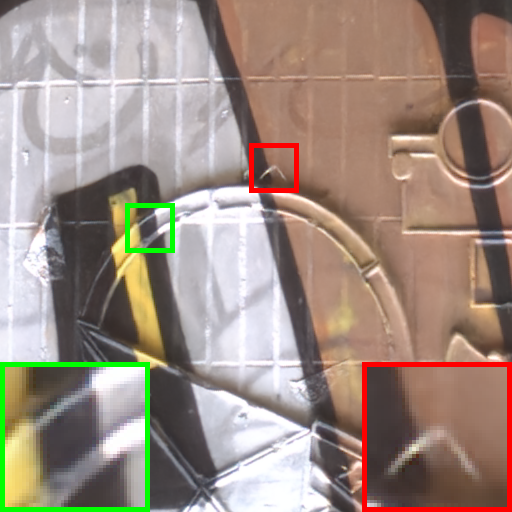}
  \centerline{\scriptsize 35.03/0.946}
  \centerline{\scriptsize PT-MWRN*}

\end{minipage}%
}
\vspace{-1ex}
\caption{\small Comparison results of our proposed method with the state-of-the-arts on DND~\cite{DND} dataset.
}
\label{fig:f7}
\end{center}}
\end{figure*}

Table~\ref{tab:training set} lists the required training set for different denoising methods.
Denoisers with superscript * are the improved counterparts using our unpaired learning scheme.
We provide floating-point operations (FLOPs), model parameters and running time comparison of different denoising models in Table~\ref{tab:running time}.
Fig.~\ref{fig:DND03} $\sim$ Fig.~\ref{Fig:MIT-IP8} also present the visualized comparison of the results by incorporating Pseudo-ISP with different pre-trained denoisers from DND~\cite{DND}, SIDD~\cite{SIDD}, SIDDPlus~\cite{ntire2020}, CC15~\cite{Nam} and MIT-IP8~\cite{MIT} datasets.
Moreover, Fig.~\ref{fig:f7} visualizes the comparison results of PT-MWRN* with the state-of-the-arts on DND~\cite{DND}.
Both in terms of quantification and visualization results, our unpaired learning scheme improves various color denoisers significantly and generalizes them well on real-world photographs.

\clearpage

\begin{table*}[t]\footnotesize
\centering
  \caption{\small{The comparison training set for different denoising methods. The symbol - indicates that no training set is required.}}
    \vspace{+2ex}
    \scalebox{1.1}{
    \begin{tabular}{c c c}
    \toprule[1pt]
    Method&
    Training set&
    Blind/Non-blind\\
    \midrule
    CDnCNN-B~\cite{DnCNN}&
    Gaussian Noise Synthesis&
    Blind\\

    Gaussian Blurring&
    -&
    Blind\\

    BM3D~\cite{dabov2008image}&
    -&
    Non-blind\\

    DIP~\cite{DIP}&
    -&
    Blind\\

    Gaussian Blurring*&
    Pseudo-ISP Synthesis&
    Blind\\

    DIP*&
    Pseudo-ISP Synthesis&
    Blind\\

    CBDNet \cite{CBDNet}&
    RENOIR~\cite{RENOIR}  + CBDNet~\cite{CBDNet} Synthesis&
    Blind\\

    CBDNet*&
    Pseudo-ISP Synthesis&
    Blind\\

    GRDN \cite{kim2019grdn}&
    SIDD~\cite{SIDD} + GAN Synthesis&
    Blind\\

    RIDNet \cite{RIDNet}&
    SIDD~\cite{SIDD} + Poly~\cite{Poly} + RENOIR~\cite{RENOIR}&
    Blind\\

    DHDN \cite{park2019densely}&
    SIDD~\cite{SIDD}&
    Blind\\

    RIDNet*&
    Pseudo-ISP Synthesis&
    Blind\\

    VDN \cite{yue2019variational}&
    SIDD~\cite{SIDD}&
    Blind\\

    DANet \cite{DANet} &
    SIDD~\cite{SIDD} + Poly~\cite{Poly} + RENOIR~\cite{RENOIR} + DANet \cite{DANet} Synthesis&
    Blind\\

    CycleISP \cite{CycleISP}&
    CycleISP \cite{CycleISP} Synthesis&
    Blind\\

    DIDN \cite{yu2019deep}&
    SIDD~\cite{SIDD}&
    Blind\\

    AINDNet \cite{Kim2020Aindnet}&
    SIDD~\cite{SIDD} + Heteroscedastic Gaussian Noise Synthesis&
    Blind\\

    PT-MWRN~\cite{cao2020progressive}&
    SIDD~\cite{SIDD} + CBDNet \cite{CBDNet} Synthesis&
    Blind\\

    MIRNet~\cite{Zamir2020MIRNet} &
    SIDD~\cite{SIDD}&
    Blind\\

    PT-MWRN*&
    Pseudo-ISP Synthesis&
    Blind\\

    \bottomrule[1pt]
    \end{tabular}
    \label{tab:training set}
    } 

\end{table*}

\begin{table*}[!t]
\begin{center}
\caption{Study on the FLOPs, model parameters and running time (the dimension of the test image is $256\times256$).}
\label{tab:running time}
\vspace{-0mm}
\setlength{\tabcolsep}{3.3pt}
\scalebox{0.65}{
\begin{tabular}{c c c c c c c c c c c c}
\toprule
 Model  &GRDN \cite{kim2019grdn}&
 DHDN \cite{park2019densely} & VDN \cite{yue2019variational}&  DANet \cite{DANet}& CycleISP \cite{CycleISP} & DIDN \cite{yu2019deep}& MIRNet~\cite{Zamir2020MIRNet} & MWCNN~\cite{MWCNN}& CBDNet~\cite{CBDNet}&RIDNet~\cite{RIDNet} & PT-MWRN~\cite{cao2020progressive}  \\
\hline
FLOPs($10^9$)  &569.0&
 1019.8 & 7.9&  14.8& 184.2 & 1489.3& 600.6& 58.3 & 6.8& 98.1& 171.1  \\
Parameters($10^6$) &34.4&
 168.2 & 49.5&  9.15& 2.8 & 217.3& 31.78& 16.1 & 62.4&1.5 & 70.2  \\
Time(ms) &118.0& 151.7 & 9.3&  5.1& 75.5 & 221.7& 205.4& 22.7 & 22.3&222.5 & 58.9 \\
\bottomrule
\end{tabular}}
\end{center}\vspace{-1.4em}
\end{table*}



\begin{figure*}[t]
\begin{center}
\scalebox{0.90}{
\begin{tabular}[t]{c@{ }c@{ }c@{ }c@{ }c@{ }c} \hspace{-2mm}
\includegraphics[width=.17\textwidth]{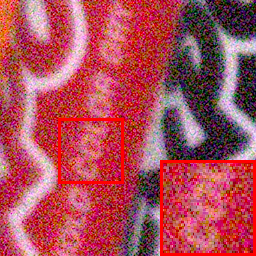}&    \hspace{-1.2mm}
\includegraphics[width=.17\textwidth]{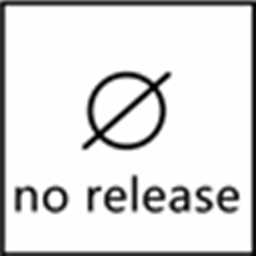}& \hspace{-1.2mm}
\includegraphics[width=.17\textwidth]{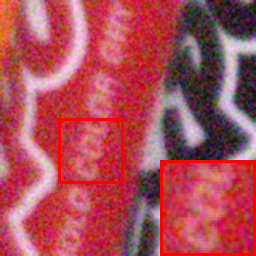}&   \hspace{-1.2mm}
\includegraphics[width=.17\textwidth]{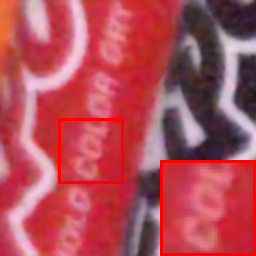}&
\hspace{-1.2mm}
\includegraphics[width=.17\textwidth]{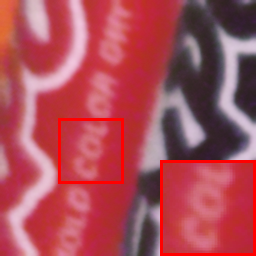}&
\hspace{-1.2mm}
\includegraphics[width=.17\textwidth]{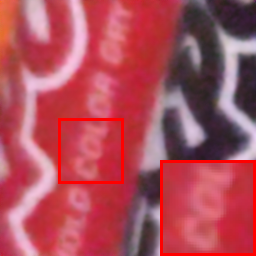}\\
Noisy&Reference&\multicolumn{2}{c}{Gaussian Blurring}&\multicolumn{2}{c}{DIP~\cite{DIP}}\\ \hspace{-2mm}
\includegraphics[width=.17\textwidth]{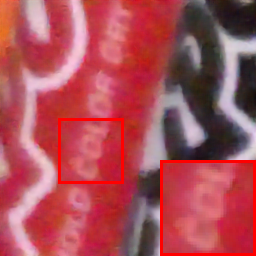}&  \hspace{-1.2mm}
\includegraphics[width=.17\textwidth]{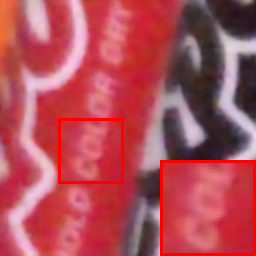}&
\hspace{-1.2mm}
\includegraphics[width=.17\textwidth]{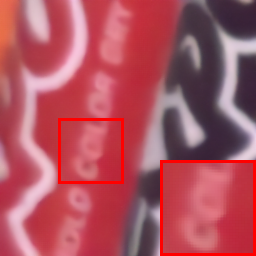}&  \hspace{-1.2mm}
\includegraphics[width=.17\textwidth]{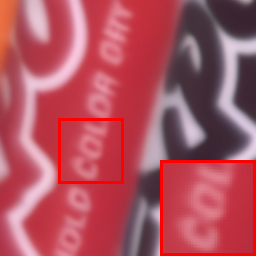}&  \hspace{-1.2mm}
\includegraphics[width=.17\textwidth]{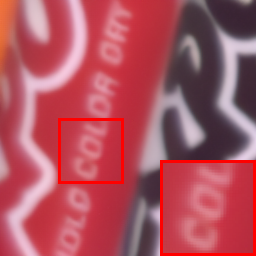}&  \hspace{-1.2mm}
\includegraphics[width=.17\textwidth]{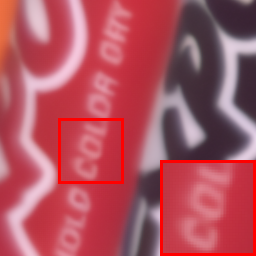}
\\
\multicolumn{2}{c}{CBDNet~\cite{CBDNet}}&\multicolumn{2}{c}{RIDNet~\cite{RIDNet}}&\multicolumn{2}{c}{PT-MWRN~\cite{cao2020progressive}}\\

\end{tabular}}
\end{center}
\vspace*{-3mm}
\caption{Denoising results of different methods on real noisy images from SIDD~\cite{SIDD}.}
\label{Fig:sidd_10_23}
\end{figure*}

\begin{figure*}[t]
\begin{center}
\scalebox{1.02}{
\begin{tabular}[t]{c@{ }c@{ }c@{ }c@{ }c@{ }c} \hspace{-2mm}
\includegraphics[width=.15\textwidth]{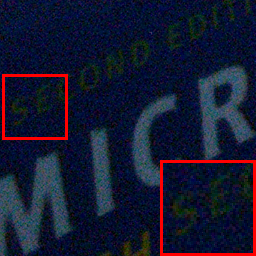}&    \hspace{-1.2mm}
\includegraphics[width=.15\textwidth]{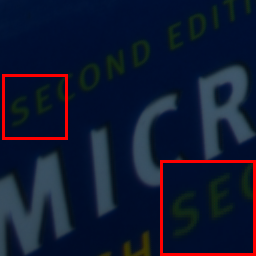}& \hspace{-1.2mm}
\includegraphics[width=.15\textwidth]{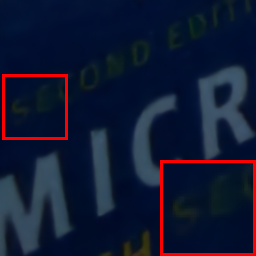}&  \hspace{-1.2mm}
\includegraphics[width=.15\textwidth]{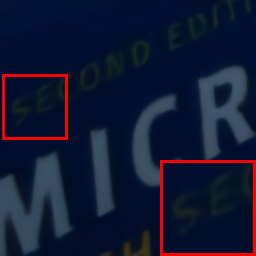}&  \hspace{-1.2mm}
\includegraphics[width=.15\textwidth]{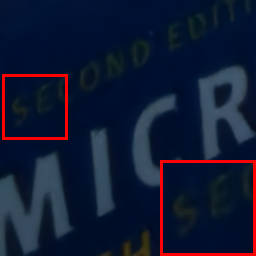}&  \hspace{-1.2mm}
\includegraphics[width=.15\textwidth]{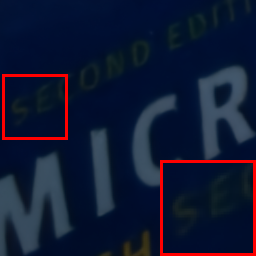}
\\
26.98/0.422 & PSNR/SSIM & \multicolumn{2}{c}{37.46/0.942 $\rightarrow$ 37.78/0.942} & \multicolumn{2}{c}{37.31/0.947 $\rightarrow$ 37.93/0.959} \\
Noisy&Reference& \multicolumn{2}{c}{RIDNet~\cite{RIDNet}} & \multicolumn{2}{c}{PT-MWRN~\cite{cao2020progressive}}
\end{tabular}}
\end{center}
\vspace*{-3mm}
\caption{Denoising results of different methods on real noisy images from SIDDPlus~\cite{ntire2020}.}
\label{Fig:siddPlus 0963}
\end{figure*}

\begin{figure*}
\begin{center}
\scalebox{0.84}{
\begin{tabular}{c@{ } c@{ }  c@{ } c@{ }  c@{ } c@{ } c}

    \multirow{4}{*}[+54.7pt]{\includegraphics[width=.31\textwidth,height=.31\textwidth ]{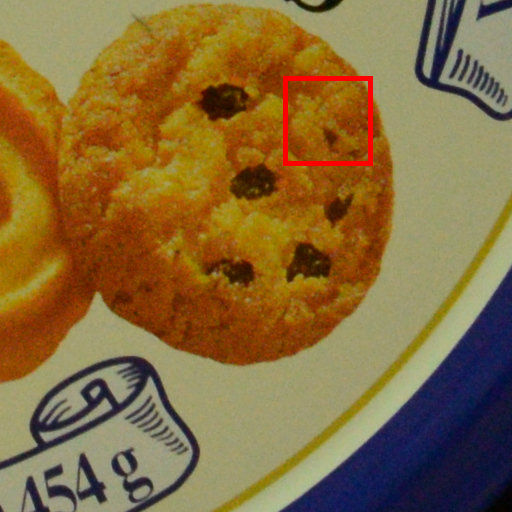}}&
    \includegraphics[width=.128\textwidth]{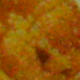}&
    \includegraphics[width=.128\textwidth]{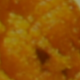}&
  	\includegraphics[width=.128\textwidth]{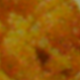}&
    \includegraphics[width=.128\textwidth]{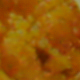}&
 	\includegraphics[width=.128\textwidth]{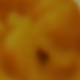}&
  	\includegraphics[width=.128\textwidth]{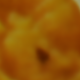}\\
     &\small33.77/0.869  & PSNR/SSIM   & \multicolumn{2}{c}{\small34.73/0.923$\rightarrow$ \small35.49/0.932} & \multicolumn{2}{c}{\small33.93/0.909$\rightarrow$  \small35.30/0.927 } \\
    & Noisy&Reference &\multicolumn{2}{c}{Gaussian Blurring} & \multicolumn{2}{c}{DIP~\cite{DIP}}\\

    &
    \includegraphics[width=.128\textwidth]{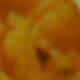}&
    \includegraphics[width=.128\textwidth]{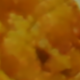}&
    \includegraphics[width=.128\textwidth]{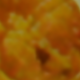}&
    \includegraphics[width=.128\textwidth]{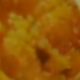}&
    \includegraphics[width=.128\textwidth]{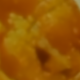}&
    \includegraphics[width=.128\textwidth]{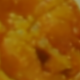}\\
        Original Noisy Image &\multicolumn{2}{c}{\small35.72/0.941 $\rightarrow$ \small36.09/0.942}  &\multicolumn{2}{c}{\small35.86/0.942  $\rightarrow$ \small36.42/0.945} & \multicolumn{2}{c}{\small36.29/0.945 $\rightarrow$ \small36.50/0.952} \\
     &\multicolumn{2}{c}{CBDNet~\cite{CBDNet}}       &\multicolumn{2}{c}{RIDNet~\cite{RIDNet}}& \multicolumn{2}{c}{PT-MWRN~\cite{cao2020progressive}}\\

\end{tabular}
}
\end{center}
\vspace*{-4mm}
\caption{Denoising results of different methods on real noisy images from CC15~\cite{Nam}.}
\label{fig:CC05}
\vspace*{-4mm}
\end{figure*}

\end{document}